\documentclass{article}

\PassOptionsToPackage{authoryear,round}{natbib}
\usepackage[preprint]{neurips_2026}

\usepackage[utf8]{inputenc} 
\usepackage[T1]{fontenc}    
\usepackage{hyperref}       
\usepackage{url}            
\usepackage{booktabs}       
\usepackage{amsfonts}       
\usepackage{nicefrac}       
\usepackage{microtype}      
\usepackage{xcolor}         

\usepackage{booktabs}
\usepackage{multirow}
\usepackage[table]{xcolor}
\usepackage{amssymb}
\usepackage{graphicx} 
\usepackage{amsmath}
\usepackage{xspace}
\usepackage{enumitem}
\usepackage{xcolor}
\usepackage{pifont}        
\usepackage[dvipsnames]{xcolor}  
\usepackage{adjustbox}
\usepackage{wrapfig}
\usepackage{cleveref}

\newcommand{\cmark}{\textcolor{ForestGreen}{\checkmark}}
\newcommand{\xmark}{\textcolor{red}{\ding{55}}}

\newcommand\method{Fruit-Fly-Foraging Algorithm\xspace}

\newcommand\me{F$^3$A\xspace}
\newcommand\mea{F$^3$A}

\setlist[itemize]{nolistsep,noitemsep,leftmargin=*}

\title{How Many Visual Tokens Do Multimodal Language Models Need? Scaling Visual Token Pruning with \mea}

\author{%
    \textbf{Yijie Huang}$^{1}$\thanks{Equal contribution.} \quad
    \textbf{Yiqun Zhang}$^{1,2}$\footnotemark[1] \quad
    \textbf{Zhuoyue Jia}$^{1}$ \quad
    \textbf{Xiaocui Yang}$^{1}$ \quad
    \textbf{Junzhao Huang}$^{1}$ \\
    \textbf{Zihan Wang}$^{1}$ \quad
    \textbf{Shi Feng}$^{1}$\thanks{Corresponding author.} \quad
    \textbf{Daling Wang}$^{1}$ \quad
    \textbf{Yifei Zhang}$^{1}$ \quad
    \textbf{Yongkang Liu}$^{3}$ \\
    $^{1}$School of Computer Science and Engineering, Northeastern University,\\
    Shenyang 110819, China \\
    $^{2}$Shanghai Artificial Intelligence Laboratory \\
    $^{3}$School of Computer and Communication Engineering, Northeastern University,\\
    Qinhuangdao 066004, China \\
    \texttt{2401837@stu.neu.edu.cn, yiqunzhang@stumail.neu.edu.cn},\\ \texttt{fengshi@cse.neu.edu.cn}
  }

\begin{document}

\maketitle

\begin{abstract}
Vision-language models improve perception by feeding increasingly long visual token sequences into language backbones, but the resulting inference cost raises a basic scaling question: as multimodal models grow, how many visual tokens are actually needed, and how should they be allocated under a fixed visual token budget? Existing training-free pruning methods typically answer this with one-shot proxies such as decoder attention, visual similarity, or conditional diversity. We argue that visual token pruning is better viewed as task-conditioned evidence search, especially under aggressive compression and across model scales. We propose \textbf{\method(\me)}, a training-free router for visual token pruning that operates before the language model consumes image tokens. \me builds lightweight question-conditioned cues, matches them to visual-grid tokens through frozen sparse sensing heads, and allocates a fixed vision token budget via coarse evidence localization, local refinement, coverage-preserving competition, and recovery of under-covered regions.
It requires no model training, no extra LLM forward pass and preserves the original multimodal prompting and decoding pipeline.
We evaluate \me on Qwen3-VL models spanning 2B to 235B parameters across dense and MoE model, covering 11 multimodal benchmarks and three retention ratios. 
Across all \textbf{10 models and 3 retention ratio settings}, \me achieves the highest compressed accuracy among FastV, VisionZip, DivPrune, and CDPruner.
At 20\% visual token retention, \me retains 93.86\% of full-token performance on average. 
When targeting 97\% of full-token performance, \me requires \textbf{only 39.9\% visual tokens} on average, compared with 50.1\% for the strongest competing baseline.
Our results suggest that scalable multimodal inference depends not only on model size, but also on search-structured allocation of task-relevant visual evidence.
\end{abstract}

\begin{figure}[htbp]
    \centering
    \includegraphics[width=\linewidth]{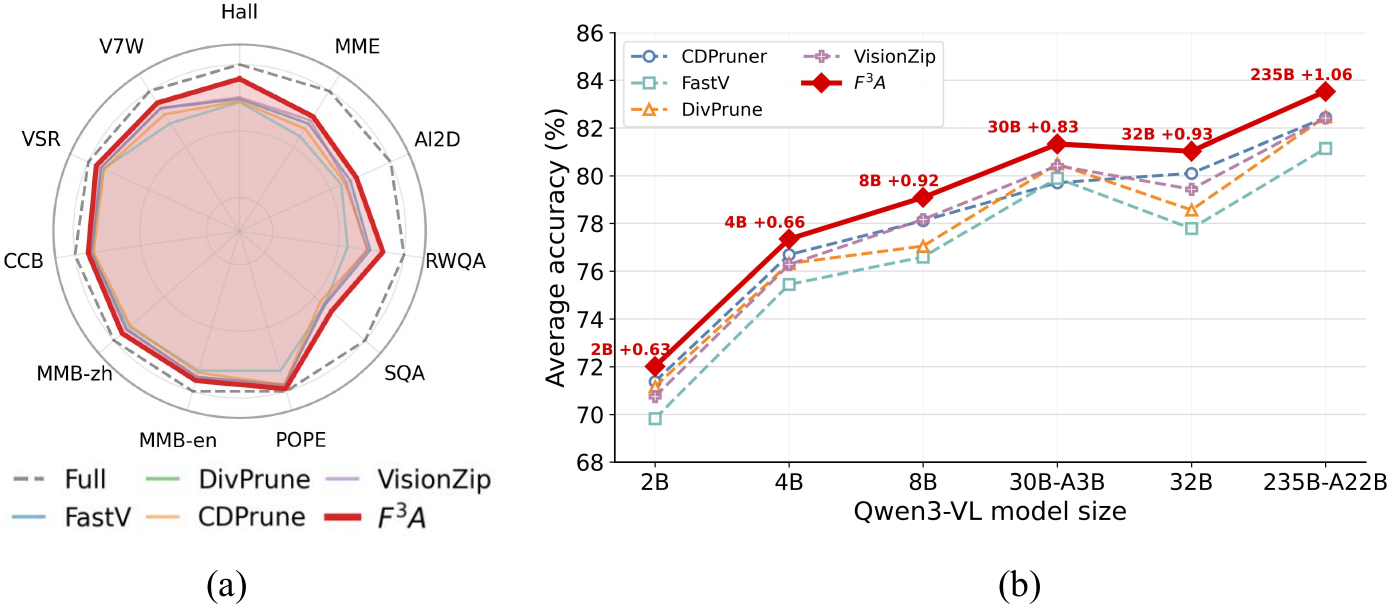}
    \caption{Compression-aware scaling on Qwen3-VL.
  (a) Average per-benchmark performance shows that \me retains stronger visual evidence across
  tasks.
  (b) Average accuracy over three retention ratios improves with model scale, \me remains
  consistently ahead of training-free pruning methods.
  }
  \label{fig:teaser-radar-scaling}
\end{figure}

\section{Introduction}

How many visual tokens does a multimodal language model actually need? This question is becoming central as vision-language models scale. Recent high-resolution and native multimodal models often improve fine-grained perception by preserving longer or more adaptive visual token sequences, enabling stronger OCR, document, chart, multi-image, and video understanding ~\citep{li2024llavaonevision,xu2024llavauhd,bai2025qwen3,bai2025qwen25vltechnicalreport}. Yet these tokens are costly: visual prefixes can be much longer than text prompts, dominate prefill computation, enlarge KV caches, and increase end-to-end latency~\citep{yang2024visionzip,Li2024InferenceOV}. As multimodal models grow from dense models to huge MoE models, visual token pruning becomes a resource-allocation problem: under a fixed visual token budget, which evidence should be kept, and how many tokens are needed to preserve full-token behavior?
This question is not answered by standard pruning evaluations. Most training-free visual token pruning methods are evaluated on one or a small number of backbone sizes, at preselected retention ratios such as 20\%, 40\%, or 60\%. Such results show whether a method works at a fixed compression point, but not how token demand changes as the multimodal model scales. In language-model scaling, the key lesson is that performance is governed by how resources are allocated, not simply by increasing one axis in isolation~\citep{kaplan2020scaling,hoffmann2022chinchilla}. Recent MLLM inference-scaling work makes the analogous point, where language-model size and visual token count jointly determine the compute--accuracy trade-off~\citep{Li2024InferenceOV}. What remains missing is a systematic study of training-free visual token pruning across a broad native multimodal model family, together with a pruning rule designed for this cross-scale allocation setting.

Existing training-free pruning methods rely on one-shot proxy signals. FastV keeps visual tokens via decoder attention~\citep{chen2024fastv}; VisionZip removes redundant tokens~\citep{yang2024visionzip}; DivPrune selects diverse tokens ~\citep{alvar2025divprune}; and CDPruner maximizes instruction-conditioned diversity~\citep{zhang2025cdpruner}. These methods are practical but treat pruning as static ranking or subset selection: computing a score (importance, redundancy, or diversity) and keeping the top subset. Under aggressive compression, this view is incomplete. A token’s value depends not only on salience or redundancy, but on its evidence for the current query. The same image should allocate tokens differently for a sign, spatial relation, chart value, verification, or small peripheral detail.

We therefore view visual token pruning as \emph{task-conditioned evidence search}. It should not only rank tokens, but also decide how a limited visual budget should move across an image-conditioned evidence landscape. It should first locate promising regions from the question, refine local evidence, avoid over-selecting redundant neighboring patches, and recover under-covered regions. We instantiate this view with \textbf{\method (\me)}, a training-free router for visual token pruning. \me operates after the vision tower and before the language backbone consumes image tokens. It builds lightweight question-conditioned cues, matches them to the visual grid through frozen sparse sensing heads, and allocates a fixed visual token budget through coarse evidence localization, local refinement, coverage-preserving competition, and recovery of under-covered regions.

We evaluate \me on Qwen3-VL models spanning 2B dense checkpoints to 235B-A22B MoE, covering eleven multimodal benchmarks and three visual token retention ratios. Figure~\ref{fig:teaser-radar-scaling} (a) shows that \me brings broad gains across benchmarks, while Figure~\ref{fig:teaser-radar-scaling} (b) shows that its average compressed accuracy remains consistently higher as the backbone scales. 
This advantage holds for every model-retention setting: in the gain heatmap of Figure~\ref{fig:qwen3-scaling} (b), all 18 cells are positive, and the largest-model 20\% retention setting still shows a +1.27-point gain over the best competing baseline. The result persists at the largest scale, on Qwen3-VL-235B-A22B, \me retains 93.86\% of full-token performance even after removing 80\% of visual tokens (Table~\ref{tab:qwen3-qwen3vl235b-results}).
We also evaluate token demand from a fixed-fidelity perspective. To recover 97\% of full-token performance, \me requires only 39.9\% visual tokens on average, while the strongest competing baseline requires 50.1\% under the same criterion (Figure \ref{fig:fixed-fidelity-97}). On Qwen3-VL-235B-A22B, \me reaches this 97\% target with 41.2\% visual tokens. This fixed-fidelity view shows that the answer is not a universal retention ratio, but a scale-dependent allocation problem shaped by how effectively the pruner preserves task evidence. Across 30 model--retention settings, \me outperforms the strongest competing baseline in every pair,
  with a two-sided sign-test $p=1.9\times10^{-9}$ (Table~\ref{tab:significance-test}).

The mechanism and deployment results support this interpretation. Ablations on Qwen3-VL-8B show that question-conditioned cues, multi-cue routing, local lock-on, and recovery of under-covered regions all matter, with larger drops under 40\% and 20\% retention (Table \ref{tab:main-ablation}). Efficiency measurements show that \me improves the accuracy--latency trade-off: at 20\% retention on Qwen3-VL-8B, it reduces end-to-end latency from 354.7 ms to 274.3 ms and reduces KV-cache footprint from 117.7 MB to 29.0 MB while remaining the most accurate pruned method at the same retention ratio (Table \ref{tab:qwen3-efficiency}). The same routing principle also transfers beyond the main Qwen3-VL family, with additional results on Qwen2.5-VL and InternVL3.5 backbones (Tables~\ref{tab:qwen25-7b-results}, \ref{tab:qwen25-32b-results}, \ref{tab:internvl35-8b-results}, \ref{tab:internvl35-38b-results}). Our contributions are:
\begin{itemize}
    \item We reframe training-free visual token pruning as a cross-scale token-allocation problem and evaluate how many visual tokens are needed to preserve full-token behavior across dense and MoE multimodal models.
    \item We propose \me, a single-pass, training-free router that performs task-conditioned evidence search before the language backbone consumes visual tokens.
    \item We provide a large-scale evaluation from Qwen3-VL-2B to Qwen3-VL-235B-A22B, showing that \me wins in all model-retention settings, requires fewer tokens to reach 97\% full-token performance, and improves the practical accuracy--efficiency frontier.
\end{itemize}

\section{Related Work}

\paragraph{Visual Token Pruning for MLLMs.}
Training-free visual token pruning reduces the long visual prefixes that make Multimodal Large Language Models (MLLMs) inference expensive. Early attention-based methods such as FastV~\citep{chen2024fastv} prune image tokens using decoder attention after early language-model layers. Other approaches avoid part of this dependence by exploiting redundancy or diversity in the visual token set: VisionZip~\citep{yang2024visionzip} compresses redundant visual tokens, DivPrune~\citep{alvar2025divprune} selects visually diverse tokens, and CDPruner~\citep{zhang2025cdpruner} further conditions diversity on the instruction. Recent training-free routers refine this design space in complementary ways. ToDRE~\citep{li2025todre} separates token diversity from task relevance and combines encoder-side selection with decoder-stage token removal. TrimTokenator~\citep{Zhang2025TrimTokenatorTA} preserves cross-modal alignment with a mutual-information criterion and then removes intra-modal redundancy through diversity-based selection. ZOO-Prune~\citep{Kim2025TrainingFreeTP} estimates token sensitivity through zeroth-order perturbations at the projection layer and combines this signal with diversity-aware selection.  
However, they still primarily instantiate pruning as proxy-driven ranking, subset selection, or staged compression.  
\me instead treats pruning as task-conditioned evidence search before LLM prefill, using question-conditioned cues and coverage-aware recovery to allocate the retained-token budget across the visual grid.

\paragraph{Bio-inspired Search and LLM Adaptation.}
Bio-inspired and population-based search has recently been used with large language models, but mainly for optimizing prompts, adapting model weights, or evolving model populations. EvoPrompt~\citep{Guo2023EvoPromptCL} and PromptBreeder~\citep{Fernando2023PromptbreederSS} evolve populations of natural-language prompts over multiple evaluation rounds; Model Swarms~\citep{Feng2024ModelSC} adapts a pool of LLM experts by collaborative movement in weight space; and GENOME~\citep{zhang2025genome} treats LLMs as an evolving population with crossover, mutation, selection, succession, and ensemble operations. \me targets a different problem. It does not optimize prompts, update model weights, compose experts, or run iterative LLM evaluations. Instead, it instantiates a foraging-style prior as a single-pass, inference-time visual token pruning. To our knowledge, \me is the first to apply bio-inspired task-conditioned search to visual token pruning.

\begin{figure}[!htbp]
    \centering
    \includegraphics[width=\linewidth]{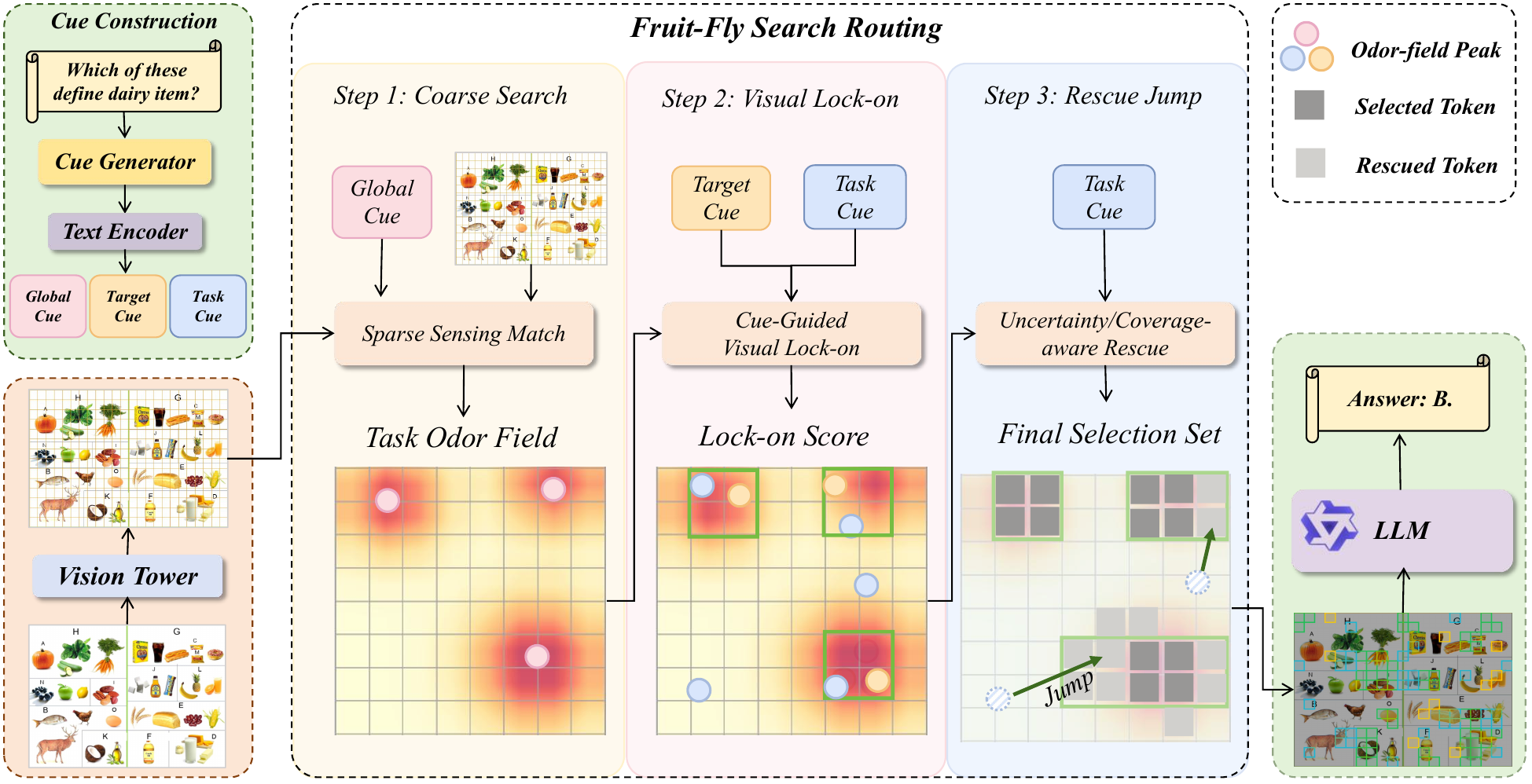}
\caption{
      Overview of \me. Prompt-conditioned cues guide a three-stage foraging process: coarse search, visual lock-on, and rescue jump.
  The selected visual tokens replace the full visual block before frozen LLM prefill, without finetuning or decoding changes.
      }
    \label{fig:method-overview}
\end{figure}

\section{Method}

\subsection{From Fruit-Fly Foraging to Token Selection}

Visual-token pruning is a fixed-budget evidence selection problem rather than ordinary saliency ranking~\citep{zhang2025cdpruner}. The usefulness of a token depends on the prompt and on what other tokens have already been kept: OCR text, peripheral objects, spatial relations, and counter-evidence may all be low-saliency but answer-critical. A one-shot score can therefore over-concentrate on obvious regions and miss distributed evidence. We instead view pruning as task-conditioned search: first obtain a cheap global evidence field, then verify local candidates, and finally reserve budget for missed or under-covered regions.

\me borrows this search order from fruit-fly optimization algorithms~\citep{Pan2012ANF,Huang2026ObscureBE}, which combine coarse exploration with local refinement. The visual grid $\Omega=\{1,\ldots,N\}$ is the search space, prompt-derived cues $\mathcal{C}(x)$ define the search target, the odor field $a_i$ is a cheap task-conditioned evidence score, the selected set $S_t$ records covered evidence, and the budget $K$ limits the search. Under this view, \me proceeds by three operators:
\begin{align}
  \mathcal{P}_1,S_1 &= \Phi_{\mathrm{coarse}}(\Omega,\mathcal{C},K), \\
  \mathcal{P}_2,S_2 &= \Phi_{\mathrm{lock}}(\mathcal{P}_1,S_1,\mathcal{C},K), \\
  S &= \Phi_{\mathrm{jump}}(\Omega,\mathcal{P}_2,S_2,\mathcal{C},K),
  \label{eq:f3-update}
\end{align}
where $\mathcal{P}_1$ and $S_1$ are the coarse candidate pool and scaffold subset, $\mathcal{P}_2$
  and $S_2$ are the locally refined candidate pool and locked-on subset, and $S$ is the final token
  set with $|S|=K$. Thus, the first stage explores promising regions, the second confirms local
  evidence and suppresses redundancy, and the third recovers uncertain or under-covered evidence.
  This decomposition is the main difference from one-shot token scoring. Figure~\ref{fig:method-overview}
  illustrates the pipeline.

 \subsection{Building the Task Odor Field}

  The first component of this search is a low-cost global evidence map. We call it the task odor field
  and define it as a scalar map $a=\{a_i\}_{i=1}^{N}$ over visual tokens, where larger $a_i$ indicates
  stronger prompt-relevant evidence. We build it by first converting the prompt into odor cues and then
  estimating cue-token responses with sparse sensing heads.

  \textbf{Odor cues.} We construct lightweight evidence queries using deterministic templates and the
  frozen tokenizer/embedding layer of the base MLLM, without external parsers, LLM extraction, training
  data, or labels. For a template string $\tau$, let $E(\tau)$ be its mean-pooled text embedding. As
  shown in Figure~\ref{fig:method-overview}, we instantiate three cue types: a global cue $c_{\mathrm{g}}$, a
  target cue $c_{\mathrm{t}}$, and a task cue $c_{\mathrm{s}}$. For open-ended prompts, $\mathcal{C}
  (x)=\{c_{\mathrm{g}},c_{\mathrm{t}},c_{\mathrm{s}}\}$ after removing unavailable cues, where
  $c_{\mathrm{g}}=\frac{1}{2}(E(x)+E(\tau_{\mathrm{global}}(x)))$ encodes the full-question context,
  $c_{\mathrm{t}}=E(\tau_{\mathrm{target}}(\tilde{x}))$ encodes the lightweight target phrase or
  queried entity, and $c_{\mathrm{s}}$ encodes task templates such as OCR/detail, counting, spatial
  relation, or verification.

  \textbf{Odor-field estimator.} Given cues, we estimate $a_i$ with frozen sparse random projections,
  not learned attention heads. Let $A_v\in\mathbb{R}^{d_s\times d_v}$ and $A_t\in\mathbb{R}^{d_s\times
  d_t}$ project visual and text features into a shared sensing space, and let $b_h$ be a sparse mask
  for head $h$. These matrices are initialized once and kept frozen. The head response and final odor
  value are
  \begin{equation}
    z_{ihc}=\left\langle
    \operatorname{norm}\!\big(b_h\odot A_vv_i\big),
    \operatorname{norm}\!\big(b_h\odot A_tc\big)
    \right\rangle,
    \qquad
    a_i=\max_{c\in\mathcal{C}(x)}\sum_{h\in\mathcal{H}_c}\omega_h(c)z_{ihc},
    \label{eq:odor-field}
  \end{equation}
  where $\operatorname{norm}(x)=x/\|x\|_2$, $\mathcal{H}_c$ contains the top-$k_h$ heads activated by
  cue $c$, and $\omega_h(c)$ is a softmax weight over these active heads. Thus, $a_i$ is not generic
  visual saliency; it is a prompt-conditioned evidence score computed before the LLM consumes the
  visual sequence.

\subsection{\method(\me)}

Given the odor field, \me allocates the token budget with a three-stage search: coarse exploration, local exploitation, and rescue exploration.

\textbf{Step 1: Coarse search.} Because odor is useful but noisy, \me first selects regions instead of committing to isolated tokens. Let $p_i=(r_i,c_i)$ be the grid coordinate of token $i$. We partition the grid into non-overlapping $w\times w$ windows and score each window by average odor:
\begin{equation}
  A(W)=\frac{1}{|W|}\sum_{i\in W}a_i,
  \qquad
  \mathcal{P}_1=\bigcup_{W\in\operatorname{TopM}(A)}W.
  \label{eq:coarse-search}
\end{equation}
Scaffold tokens $S_1=\operatorname{Scaffold}(\mathcal{P}_1)$ are kept from the selected windows, giving the next stage a spatially covered candidate pool.

\textbf{Step 2: Visual lock-on.} Within the coarse pool, \me confirms local evidence and suppresses repeated selections. For neighborhood $\mathcal{N}_r(i)=\{j:\|p_i-p_j\|_\infty\le r\}$, we estimate local support as
\begin{equation}
  \ell_i=
  \frac{1}{2|\mathcal{N}_r(i)|}\sum_{j\in\mathcal{N}_r(i)}a_j
  +\frac{1}{2}\max_{j\in\mathcal{N}_r(i)}s_j,
  \label{eq:local-support}
\end{equation}
where $s_j$ is the normalized task score from cue agreement, option support when present, and local detail contrast. Redundancy is measured by visual similarity and spatial proximity,
\begin{equation}
  \kappa(p_i,p_j)=\exp\!\left(-\frac{\|p_i-p_j\|_2^2}{2\sigma_p^2}\right),
  \quad
  r_i=\max_{j\in S_t}\left[\operatorname{sim}(\bar v_i,\bar v_j)+\kappa(p_i,p_j)\right],
  \label{eq:redundancy}
\end{equation}
where $\bar v_i$ is the $\ell_2$-normalized visual token. The lock-on score is
\begin{equation}
  m_i=a_i+\lambda\ell_i-\beta r_i,
  \qquad i\in\mathcal{P}_1.
  \label{eq:lock-score}
\end{equation}
This implements an inhibition-of-return effect: locally supported tokens are favored, while redundant neighboring patches are discouraged.

\textbf{Step 3: Rescue jump.} To avoid missing small objects, peripheral text, or counter-evidence outside the selected windows, \me reserves a fraction $\alpha_{\mathrm{jump}}$ of the budget for rescue. For multiple-choice prompts, uncertainty is the normalized margin between the top two option-support scores,
\begin{equation}
  u_i=1-\nu\!\left(h_{i,(1)}-h_{i,(2)}\right),
  \label{eq:uncertainty-mc}
\end{equation}
and for open-ended prompts we use $u_i=1-\nu(g_i)$, where $g_i$ is global-cue agreement. Coverage by the current subset is
\begin{equation}
  \operatorname{cov}(i,S)=
  \max_{j\in S}
  \left[\alpha_c\operatorname{sim}(\bar v_i,\bar v_j)+(1-\alpha_c)\kappa(p_i,p_j)\right].
  \label{eq:coverage}
\end{equation}
The rescue score
\begin{equation}
  q_i=a_i+\gamma u_i-\eta\operatorname{cov}(i,S_2),
  \qquad i\in\Omega\setminus S_2
  \label{eq:rescue-score}
\end{equation}
selects tokens that are still task-relevant but insufficiently represented by the current subset. Component sensitivity is studied in the main ablation table and Appendix~\ref{app:ablation}.

\textbf{Sequence reconstruction.} After selecting $S$, \me replaces the full visual block with $V_S$ and leaves the rest of the inference pipeline unchanged. The original text prefix, prompt template, attention mask, decoding configuration, and model weights are preserved. For grid-aware MLLMs, selected token indices are mapped back to their original grid coordinates before recomputing position ids. Thus, \me only shortens the visual prefix seen by the language model and requires no finetuning, calibration data, extra LLM forward pass, or fallback to full-token inference.

\begin{table*}[!htbp]
  \caption{Qwen3-VL scaling results for Qwen3-VL-2B. \textbf{Acc.}\ is the average accuracy over the non-MME datasets with a full-token baseline shown in the corresponding table, and \textbf{Rel.}\ is the ratio between this average accuracy and the corresponding full-token result.}
  \label{tab:qwen3-qwen3vl2b-results}
  \centering
  \scriptsize
  \setlength{\tabcolsep}{1.2pt}
  \renewcommand{\arraystretch}{0.96}
  \begin{tabular*}{\textwidth}{@{\extracolsep{\fill}}llrrrrrrrrrrr|cc@{}}
  \toprule
  Ratio & Method & Hall & MME & AI2D & RWQA & SQA & POPE & MB$^{\mathrm{en}}$ & MB$^{\mathrm{zh}}$ & CCB & VSR & V7W & Acc & Rel \\
  \midrule
  100\% & Qwen3-VL-2B & 52.41 & 1983.3 & 74.26 & 65.10 & 86.25 & 87.30 & 77.50 & 75.93 & 68.88 & 75.20 & 86.10 & 74.89 & 100.00 \\
  \midrule
  \multirow{5}{*}{60\%} & CDPruner & 51.36 & 1919.2 & \textbf{71.34} & 62.88 & \textbf{74.40} & 87.12 & 77.08 & 75.33 & 68.23 & 74.14 & \textbf{86.06} & 72.79 & 97.20 \\
   & FastV & 52.52 & 1934.5 & 70.92 & 63.01 & 71.99 & 86.98 & 76.00 & 74.19 & 67.45 & 74.22 & 85.84 & 72.31 & 96.55 \\
   & DivPrune & 51.78 & 1936.3 & 71.18 & 62.35 & 74.29 & \textbf{87.32} & \textbf{77.50} & \textbf{75.51} & 66.99 & 74.22 & 85.40 & 72.65 & 97.01 \\
   & VisionZip & 52.41 & \textbf{1961.3} & 69.92 & 62.35 & 73.73 & 87.31 & \textbf{77.50} & 75.44 & 67.82 & \textbf{74.39} & 86.04 & 72.69 & 97.06 \\
   & \textbf{\me (Ours)} & \textbf{52.83} & 1947.4 & 71.02 & \textbf{63.68} & 73.52 & 87.31 & \textbf{77.50} & 75.49 & \textbf{68.32} & \textbf{74.39} & 85.76 & \textbf{72.98} & \textbf{97.45} \\
  \midrule
  \multirow{5}{*}{40\%} & CDPruner & 50.42 & 1895.8 & 69.20 & 62.35 & 73.37 & 87.00 & 77.02 & 74.79 & 65.93 & 73.00 & 84.76 & 71.78 & 95.85 \\
   & FastV & \textbf{52.10} & 1885.1 & 67.68 & 61.05 & 72.34 & 85.77 & 74.66 & 73.20 & \textbf{66.80} & 72.67 & 83.16 & 70.94 & 94.73 \\
   & DivPrune & 51.57 & 1885.0 & \textbf{69.27} & 61.57 & \textbf{73.88} & 87.13 & 76.83 & 73.94 & 65.52 & 72.91 & 84.06 & 71.67 & 95.69 \\
   & VisionZip & 51.68 & \textbf{1899.9} & 68.13 & 61.83 & 73.83 & 86.90 & 76.46 & 74.33 & 66.30 & 72.59 & 84.84 & 71.69 & 95.72 \\
   & \textbf{\me (Ours)} & 51.47 & 1899.8 & 69.20 & \textbf{62.75} & 73.22 & \textbf{87.31} & \textbf{77.73} & \textbf{75.58} & 66.71 & \textbf{74.30} & \textbf{85.52} & \textbf{72.38} & \textbf{96.64} \\
  \midrule
  \multirow{5}{*}{20\%} & CDPruner & 48.73 & 1767.8 & 65.28 & 59.35 & 72.09 & 85.48 & 75.93 & 72.41 & 63.17 & 70.70 & 82.34 & 69.55 & 92.86 \\
   & FastV & 48.73 & 1705.1 & 63.44 & 56.08 & 71.63 & 78.37 & 72.07 & 70.71 & 62.57 & 63.01 & 75.58 & 66.22 & 88.42 \\
   & DivPrune & 49.15 & 1774.1 & \textbf{65.58} & 59.87 & 71.99 & 85.16 & 74.47 & 71.19 & 62.06 & \textbf{71.03} & 81.20 & 69.17 & 92.36 \\
   & VisionZip & 47.16 & 1621.2 & 62.08 & 58.56 & 71.16 & 82.50 & 74.50 & 70.87 & 63.58 & 69.07 & 78.98 & 67.85 & 90.59 \\
   & \textbf{\me (Ours)} & \textbf{49.89} & \textbf{1802.6} & \textbf{65.58} & \textbf{61.57} & \textbf{73.11} & \textbf{86.94} & \textbf{76.07} & \textbf{73.87} & \textbf{65.01} & \textbf{71.03} & \textbf{83.50} & \textbf{70.66} & \textbf{94.34} \\
  \bottomrule
  \end{tabular*}
\end{table*}

\section{Experiments}
\subsection{Experimental setup}


\textbf{Models.}
Our main study uses Qwen3-VL-Instruct~\citep{bai2025qwen3} at: 2B, 4B, 8B, 30B-A3B,
32B, and 235B-A22B, covering both dense and MoE backbones. For each model, we compare full-token inference with 60\%, 40\%, and 20\% visual token retention. To test cross-family transfer, we evaluate Qwen2.5-VL-7B/32B~\citep{bai2025qwen25vltechnicalreport} and InternVL3.5-8B/38B~\citep{chen2024internvl}.

\textbf{Evaluation benchmarks.}
We evaluate on eleven multimodal benchmarks: HallusionBench~\citep{guan2024hallusionbench}, MME~\citep{fu2023mme}, AI2D~\citep{kembhavi2016ai2d}, RealWorldQA~\citep{xai2024realworldqa}, ScienceQA-IMG~\citep{lu2022scienceqa}, POPE~\citep{li2023pope}, MMBench-en, MMBench-CN, CCBench~\citep{liu2024mmbench},  VSR~\citep{liu2023vsr}, and Visual7W~\citep{zhu2016visual7w}. MME is reported using the official MME score, while other datasets use accuracy (\%). Dataset abbreviations and details are provided in Appendix~\ref{app:datasets}, Table~\ref{tab:dataset}.

\textbf{Baselines.}
Following prior training-free visual-token pruning
studies, we compare with FastV~\citep{chen2024fastv},
DivPrune~\citep{alvar2025divprune}, CDPruner~\citep{zhang2025cdpruner}, and VisionZip~\citep{yang2024visionzip}. For each model and benchmark, all methods use the same prompt template,
decoding configuration, split, metric, and retention ratios. Pruning only changes the number of visual
tokens passed to the multimodal LLM prefill; model weights, prompts, and answer post-processing remain
unchanged. No method uses task-specific finetuning, calibration examples, or benchmark labels, and all
reported results are averaged over three repeated runs. For \me, the same hyperparameters are used for all datasets, backbones, and retention ratios; their values are listed in Appendix~\ref{app:hyperparams}, with
stability analysis in Table~\ref{tab:hyperparam-stability}. All evaluations are conducted on 8 $\times$ H200 GPUs.

\begin{table*}[!htbp]
  \caption{Qwen3-VL scaling results for Qwen3-VL-235B-A22B. \textbf{Acc.}\ is the average accuracy over the non-MME datasets with a full-token baseline shown in the corresponding table, and \textbf{Rel.}\ is the ratio between this average accuracy and the corresponding full-token result.} 
  \label{tab:qwen3-qwen3vl235b-results}
  \centering
  \scriptsize
  \setlength{\tabcolsep}{1.2pt}
  \renewcommand{\arraystretch}{0.96}
  \begin{tabular*}{\textwidth}{@{\extracolsep{\fill}}llrrrrrrrrrrr|cc@{}}
  \toprule
  Ratio & Method & Hall & MME & AI2D & RWQA & SQA & POPE & MB$^{\mathrm{en}}$ & MB$^{\mathrm{zh}}$ & CCB & VSR & V7W & Acc & Rel \\
  \midrule
  100\% & Qwen3-VL-235B & 64.19 & {2631.7} & 88.50 & 76.99 & 98.00 & 90.51 & {88.79} & {88.10} & {86.77} & {92.61} & {91.73} & 86.62 & 100.00 \\
  \midrule

  \multirow{5}{*}{60\%}
  & CDPruner 
  & 60.94 & 2548.2 & 85.41 & 75.82 & 93.74 & 89.68 
  & 86.94 & 86.01 & 85.62 & 92.05 & 89.42 
  & 84.56 & 97.63 \\

  & FastV 
  & 61.27 & 2511.6 & 84.90 & 74.98 & 93.21 & 88.94 
  & 86.33 & 85.42 & 85.08 & 92.11 & 87.73 
  & 84.00 & 96.97 \\

  & DivPrune 
  & 61.45 & 2562.4 & 85.96 & 75.94 & 94.02 & 89.77 
  & 87.20 & 86.38 & 85.91 & 92.18 & 88.61 
  & 84.74 & 97.83 \\

  & VisionZip 
  & 62.12 & \textbf{2588.9} & 85.72 & 75.66 & 94.10 & 89.82 
  & \textbf{87.42} & 86.71 & 85.74 & 92.33 & 88.52 
  & 84.81 & 97.92 \\

  & \textbf{\me (Ours)} 
  & \textbf{62.98} & 2580.3 & \textbf{86.43} & \textbf{76.86} & \textbf{94.66} & \textbf{90.10} 
  & 87.25 & \textbf{86.76} & \textbf{86.37} & \textbf{92.39} & \textbf{90.06} 
  & \textbf{85.39} & \textbf{98.58} \\

  \midrule

  \multirow{5}{*}{40\%}
  & CDPruner 
  & 58.32 & 2412.6 & 82.02 & 74.11 & 91.52 & 88.74 
  & 85.84 & 85.12 & 84.48 & 90.02 & 87.36 
  & 82.75 & 95.54 \\

  & FastV 
  & 57.33 & 2366.5 & 82.41 & 72.98 & 90.96 & 87.53 
  & 85.02 & 84.66 & 84.01 & 89.98 & 87.12 
  & 82.20 & 94.90 \\

  & DivPrune 
  & 57.96 & 2428.4 & 83.38 & 74.02 & 91.73 & 88.92 
  & 86.01 & 85.36 & 84.92 & 89.12 & 87.60 
  & 82.90 & 95.71 \\

  & VisionZip 
  & 59.02 & \textbf{2440.1} & \textbf{83.44} & 74.21 & 90.88 & \textbf{89.04} 
  & 86.33 & 85.70 & 84.01 & 90.27 & 88.02 
  & 83.09 & 95.93 \\

  & \textbf{\me (Ours)} 
  & \textbf{59.45} & 2439.7 & 83.42 & \textbf{75.21} & \textbf{92.71} & 89.96 
  & \textbf{86.34} & \textbf{86.14} & \textbf{85.22} & \textbf{91.41} & \textbf{89.44} 
  & \textbf{83.93} & \textbf{96.90} \\

  \midrule

  \multirow{5}{*}{20\%}
  & CDPruner 
  & 54.01 & 2268.4 & 77.32 & 70.85 & 87.62 & 87.31 
  & 83.72 & 82.96 & 82.11 & 87.94 & 86.22 
  & 80.01 & 92.37 \\

  & FastV 
  & 52.33 & 2104.7 & 76.41 & 66.72 & 85.90 & 80.22 
  & 81.55 & 80.74 & 80.88 & 86.31 & 81.45 
  & 77.25 & 89.18 \\

  & DivPrune 
  & 53.88 & 2291.6 & 77.95 & 71.42 & 87.11 & 86.84 
  & 82.96 & 82.41 & 81.94 & \textbf{88.22} & 85.98 
  & 79.87 & 92.21 \\

  & VisionZip 
  & 53.67 & 2240.5 & 77.22 & 71.03 & 86.88 & 87.02 
  & 83.01 & 82.73 & 82.36 & 87.90 & 86.74 
  & 79.86 & 92.19 \\

  & \textbf{\me (Ours)} 
  & \textbf{54.72} & \textbf{2366.1} & \textbf{78.94} & \textbf{73.20} & \textbf{89.53} & \textbf{89.03} 
  & \textbf{85.33} & \textbf{84.06} & \textbf{83.66} & 87.81 & \textbf{87.50} 
  & \textbf{81.38} & \textbf{93.86} \\

  \bottomrule
  \end{tabular*}
\end{table*}

\begin{figure*}[!htbp]
  \centering
  \includegraphics[width=\textwidth]{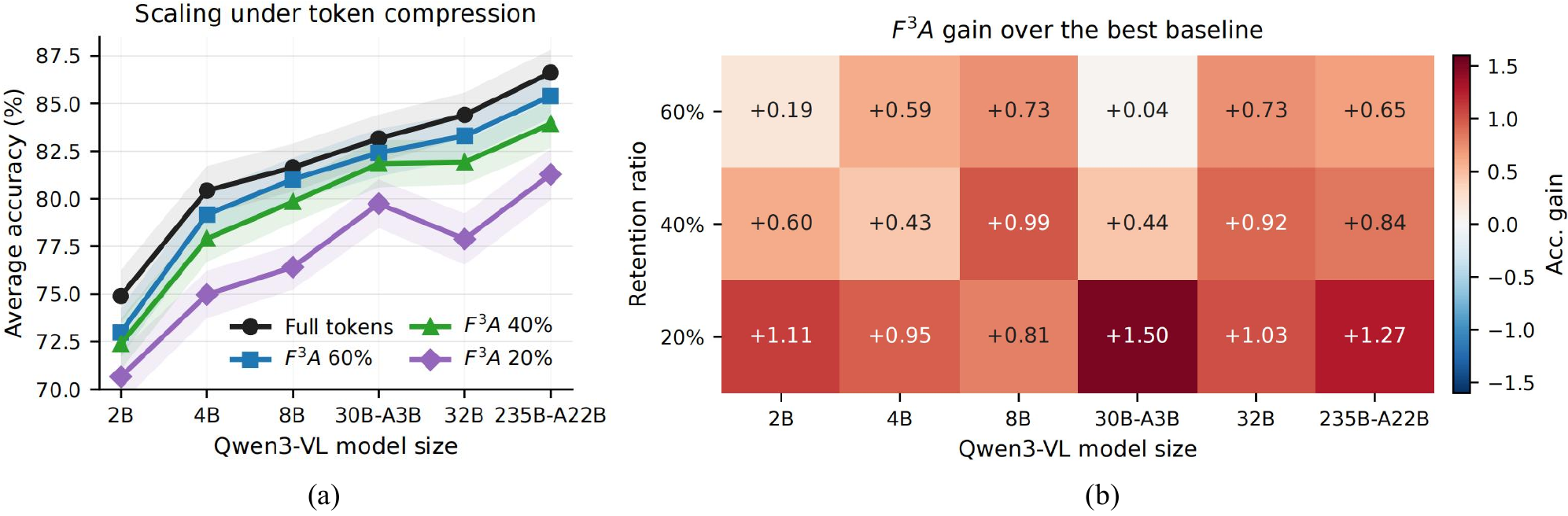}
  \caption{
  Compression-aware scaling on Qwen3-VL.
  (a) Average accuracy of full-token inference and \me at 60\%, 40\%, and 20\% retention across six
  scales.
  (b) Accuracy gain of \me over the strongest baseline at each scale and budget.
  }
  \label{fig:qwen3-scaling}
\end{figure*}

\subsection{Main Results}

  Figure~\ref{fig:qwen3-scaling} summarizes the compression-aware scaling behavior of \me on Qwen3-VL from 2B to 235B. The left panel shows that average accuracy generally improves with model scale
  under all token budgets, indicating that visual token compression does not remove the benefit of
  scaling; even at 20\% retention, the curve follows the same upward trend. The right panel compares
  \me with the strongest competing baseline at each model size and retention ratio. All cells are
  positive, showing consistent gains across scales and budgets; on Qwen3-VL-235B, \me still improves
  by +0.65, +0.84, and +1.27 points at 60\%, 40\%, and 20\% retention, respectively.Tables~\ref{tab:qwen3-qwen3vl2b-results} and~\ref{tab:qwen3-qwen3vl235b-results} report the two
  endpoint scales, while Table~\ref{tab:additional-backbone-summary} summarizes the remaining Qwen3-VL models together with Qwen2.5-VL-7B/32B and InternVL3.5-8B/38B. Full per-dataset results for these additional models are provided in Appendix~\ref{app:mainexp}. Averaged
  across Qwen3-VL model sizes, \me retains 98.58\%, 97.19\%, and 93.86\% of full-token performance
  at 60\%, 40\%, and 20\% retention. The additional Qwen2.5-VL and InternVL3.5 results further show
  that \me is not specific to Qwen3-VL: it remains competitive and is best in most settings,
  although a few cases favor VisionZip or CDPruner, suggesting that pruning behavior can depend on
  the backbone architecture. Overall, these results show that our method preserves the benefit of
  model scaling while reducing the visual sequence length processed by the MLLM, and generally
  transfers across model families. A paired significance analysis over 30 model--retention settings further confirms this consistency:
  \me beats the strongest non-\me baseline in all pairs, with $p=1.9\times10^{-9}$
  (Table~\ref{tab:significance-test}).

\paragraph{Fixed-fidelity token demand.}
The fixed-retention results above follow the standard pruning protocol: given a retention ratio, they measure the resulting accuracy. To answer the title question more directly, we also use the complementary fixed-fidelity view: given a target fidelity to the full-token model, how many visual tokens are required? For a model
$M$ and pruning method $m$, we define
\begin{equation}
r_\tau(M,m)=\min_{\rho}
\left\{
\rho:
A(M,m,\rho)/A_{\mathrm{full}}(M) \ge \tau
\right\},
\end{equation}
where $\rho$ is the visual token retention ratio and $\tau$ is the target fraction of full-token performance. We estimate $r_\tau$ by linearly interpolating the measured 20\%, 40\%, 60\%, and 100\% retention points.

\begin{figure}[!htbp]
    \centering
    \includegraphics[width=1\linewidth]{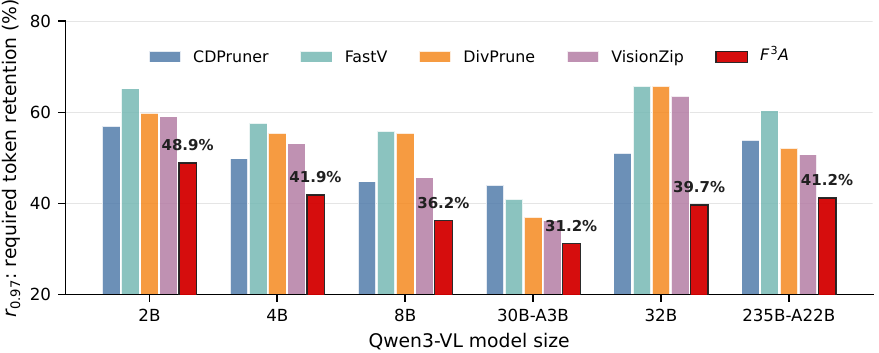}
    \caption{On Qwen3-VL family, each bar is the minimum visual token retention required to preserve 97\% of full-token performance. \me requires \textbf{fewer tokens} than all baselines across model scales.}
  \label{fig:fixed-fidelity-97}
\end{figure}

We use $\tau=0.97$ as the primary near-full fidelity target, allowing at most 3\% relative degradation from full-token inference. Figure~\ref{fig:fixed-fidelity-97} shows the resulting token demand across Qwen3-VL scales, while Appendix~\ref{app:fixed_fidelity_sensitivity} reports 95\% and 98\% sensitivity. Under this view, \me requires fewer visual tokens than every competing
baseline: it reaches 97\% full-token performance with only 39.9\% visual tokens on average, compared with 50.1\% for the strongest baseline. On Qwen3-VL-235B-A22B, \me needs 41.2\% tokens. Thus, \me not only improves accuracy at fixed retention ratios, but also lowers the token budget needed to preserve near-full model behavior.

\begin{table*}[!htbp]
\caption{Summary of additional backbone results reported outside the main endpoint tables. Entries are Acc., the average accuracy over non-MME benchmarks with a full-token baseline. Bold indicates the best pruning method at the same retention ratio and model scale.}
\centering
\small
\setlength{\tabcolsep}{4.5pt}
\renewcommand{\arraystretch}{0.96}
\begin{tabular*}{0.9\textwidth}{@{\extracolsep{\fill}}llcccccccc@{}}
\toprule
\multirow{2}{*}{Ratio} & \multirow{2}{*}{Method} & \multicolumn{4}{c}{Qwen3-VL} & \multicolumn{2}{c}{Qwen2.5-VL} & \multicolumn{2}{c}{InternVL3.5} \\
\cmidrule(lr){3-6} \cmidrule(lr){7-8} \cmidrule(lr){9-10}
 & & 4B & 8B & 30B-A3B & 32B & 7B & 32B & 8B & 38B \\
\midrule
100\% & Full & 80.43 & 81.64 & 83.16 & 84.40 & 78.76 & 81.45 & 79.05 & 82.93 \\
\midrule
\multirow{5}{*}{60\%} & CDPruner & 78.58 & 80.29 & 81.58 & 82.57 & 76.88 & 79.86 & 76.93 & 81.67 \\
 & FastV & 78.24 & 79.53 & 82.16 & 81.44 & 76.28 & 79.95 & 76.91 & 81.43 \\
 & DivPrune & 78.36 & 79.62 & 82.17 & 81.43 & 76.99 & 79.59 & 77.14 & 81.24 \\
 & VisionZip & 78.45 & 80.06 & 82.37 & 81.61 & 76.72 & 79.85 & 77.39 & 81.27 \\
 & \me (Ours) & \textbf{79.17} & \textbf{81.02} & \textbf{82.41} & \textbf{83.30} & \textbf{77.19} & \textbf{80.28} & \textbf{77.75} & \textbf{82.03} \\
\midrule
\multirow{5}{*}{40\%} & CDPruner & 77.47 & 78.84 & 80.43 & 81.00 & 74.89 & 79.37 & 75.39 & 80.40 \\
 & FastV & 76.36 & 77.88 & 80.59 & 78.45 & 74.99 & 78.90 & 75.42 & 80.25 \\
 & DivPrune & 76.85 & 77.71 & 81.09 & 79.13 & 74.93 & 78.77 & 75.98 & 79.75 \\
 & VisionZip & 77.18 & 78.85 & 81.40 & 79.86 & 75.24 & 78.58 & 76.28 & 79.97 \\
 & \me (Ours) & \textbf{77.90} & \textbf{79.84} & \textbf{81.84} & \textbf{81.92} & \textbf{75.46} & \textbf{79.69} & \textbf{76.34} & \textbf{80.80} \\
\midrule
\multirow{5}{*}{20\%} & CDPruner & 74.02 & 75.22 & 77.12 & 76.75 & 73.00 & 74.74 & 72.33 & 77.00 \\
 & FastV & 71.75 & 72.37 & 76.95 & 73.48 & 70.55 & 74.76 & 70.19 & 76.85 \\
 & DivPrune & 73.75 & 73.82 & 78.24 & 75.14 & 73.22 & 75.01 & 72.71 & 76.56 \\
 & VisionZip & 73.20 & 75.60 & 77.51 & 76.85 & 73.30 & 74.40 & 73.06 & 76.19 \\
 & \me (Ours) & \textbf{74.97} & \textbf{76.41} & \textbf{79.74} & \textbf{77.88} & \textbf{73.42} & \textbf{75.78} & \textbf{73.97} & \textbf{77.70} \\
\bottomrule
\end{tabular*}

\label{tab:additional-backbone-summary}
\end{table*}

\subsection{Ablation Study}

\begin{table*}[!htbp]
\caption{Main ablation study on Qwen3-VL-8B. We report accuracy (\%) on HallusionBench, RealWorldQA, and AI2D under 60\%, 40\%, and 20\% visual token retention. $\Delta$ denotes the average accuracy drop relative to the full \me variant at the same retention ratio.}
\centering
\small
\setlength{\tabcolsep}{4.5pt}
\begin{tabular}{llrrrr|rr}
\toprule
Ratio & Variant & Hall & RWQA & AI2D & Avg. & $\Delta$ & Rel. \\
\midrule
\multirow{5}{*}{60\%} & \textbf{Full \me} & \textbf{62.93} & \textbf{69.93} & \textbf{81.12} & \textbf{71.33} & -- & 100.00 \\
 & w/o Odor Cue    & 61.84 & 69.28 & 79.57 & 70.23 & -1.10 & 98.46 \\
 & w/o Multi-Cue   & 61.79 & 69.15 & 79.60 & 70.18 & -1.15 & 98.39 \\
 & w/o Visual Lock-on & 61.52 & 69.02 & 79.70 & 70.08 & -1.25 & 98.25 \\
 & w/o Rescue Jump & 61.68 & 69.08 & 79.63 & 70.13 & -1.20 & 98.32 \\
\midrule
\multirow{5}{*}{40\%} & \textbf{Full \me} & \textbf{62.40} & \textbf{68.24} & \textbf{78.53} & \textbf{69.72} & -- & 100.00 \\
 & w/o Odor Cue    & 58.09 & 67.58 & 76.52 & 67.40 & -2.33 & 96.66 \\
 & w/o Multi-Cue   & 58.40 & 67.45 & 76.23 & 67.36 & -2.36 & 96.61 \\
 & w/o Visual Lock-on & 57.46 & 67.58 & 76.36 & 67.13 & -2.59 & 96.29 \\
 & w/o Rescue Jump & 58.62 & 66.67 & 76.17 & 67.15 & -2.57 & 96.32 \\
\midrule
  \multirow{5}{*}{20\%} & \textbf{Full \me} & \textbf{57.67} & \textbf{64.58} & \textbf{72.80} & \textbf{65.02} & -- &
  100.00 \\
   & w/o Odor Cue    & 52.83 & 62.22 & 71.11 & 62.05 & -2.96 & 95.44 \\
   & w/o Multi-Cue   & 54.30 & 62.22 & 71.92 & 62.81 & -2.20 & 96.61 \\
   & w/o Visual Lock-on & 53.88 & 61.05 & 71.21 & 62.05 & -2.97 & 95.43 \\
   & w/o Rescue Jump & 54.51 & 61.96 & 71.73 & 62.73 & -2.28 & 96.49 \\
\bottomrule
\end{tabular}
\label{tab:main-ablation}
\end{table*}

Table~\ref{tab:main-ablation} ablates the main mechanisms in \me on Qwen3-VL-8B at 60\%, 40\%, and
20\% retention, with the normalized bar visualization provided in Appendix~\ref{app:ablation},
Figure~\ref{fig:main-ablation-bars}. Removing the text-derived odor cue reduces average accuracy
by 1.10, 2.33, and 2.96 points, respectively, confirming the need for question-conditioned
pruning. Removing multi-cue construction also degrades performance, showing that global relevance,
option-level evidence, and contrastive signals are complementary for selecting task-relevant
visual tokens. The visual stages are similarly important: disabling visual lock-on drops accuracy
by 1.25, 2.59, and 2.97 points, while removing rescue jumps drops it by 1.20, 2.57, and 2.28
points. These results support the foraging design: under a limited token budget, \me needs both
local exploitation and exploratory recovery to avoid collapsing onto narrow high-score regions.

\subsection{Efficiency Analysis}
We evaluate end-to-end efficiency on a single GPU using Qwen3-VL-8B over HallusionBench,
RealWorldQA, and POPE, reporting the average score, generation latency, KV-cache footprint, and peak extra memory under the same retention ratios as
the main experiments. We report efficiency on the 8B model, the common scale size used by the compared baselines, for a fair
  same-backbone comparison. Table~\ref{tab:qwen3-efficiency} shows that \me achieves the best
accuracy--efficiency tradeoff among pruning methods: at 60\%, 40\%, and 20\% retention, it obtains
72.96\%, 73.02\%, and 69.38\% average score, while reducing latency from 354.7 ms to 335.5 ms, 313.3 ms, and 274.3 ms, yielding 1.06$\times$, 1.13$\times$, and
1.29$\times$ speedup. Since all methods keep the same number of visual tokens at a fixed ratio,
their KV footprints are nearly identical; the key difference is whether the token selection
preserves useful evidence without excessive overhead. Compared with the strongest non-\me
score at each budget, \me improves by 1.25, 1.52, and 0.25 points, while keeping peak extra memory
low at 167.0 MB, 173.8 MB, and 171.6 MB. These results show that \me improves compressed accuracy
while preserving the practical latency and memory benefits of visual token reduction.

\begin{table*}[!htbp]
\caption{Efficiency on Qwen3-VL-8B averaged over HallusionBench, RealWorldQA, and POPE. E2E latency includes pruning and generation; speedup is relative to full-token inference. KV and Mem. denote KV-cache footprint and peak extra memory.}
\centering
\scriptsize
\setlength{\tabcolsep}{2.4pt}
\resizebox{\textwidth}{!}{%
\begin{tabular}{l*{12}{c}}
\toprule
\multirow{2.5}{*}{Method} & \multicolumn{4}{c}{Retention Ratio 0.6} & \multicolumn{4}{c}{Retention Ratio 0.4} & \multicolumn{4}{c}{Retention Ratio 0.2} \\
\cmidrule(lr){2-5} \cmidrule(lr){6-9} \cmidrule(lr){10-13}
 & Score (\%) $\uparrow$ & E2E Lat. (ms) $\downarrow$ & KV (MB) $\downarrow$ & Mem. (MB) $\downarrow$ & Score (\%) $\uparrow$ & E2E Lat. (ms) $\downarrow$ & KV (MB) $\downarrow$ & Mem. (MB) $\downarrow$ & Score (\%) $\uparrow$ & E2E Lat. (ms) $\downarrow$ & KV (MB) $\downarrow$ & Mem. (MB) $\downarrow$ \\
\midrule
Original & 73.23 & 354.7 (1.00$\times$) & 117.7 & 209.7 & 73.23 & 354.7 (1.00$\times$) & 117.7 & 209.7 & 73.23 & 354.7 (1.00$\times$) & 117.7 & 209.7 \\
\midrule
FastV & 71.36 & 343.8 (1.03$\times$) & 73.3 & 174.0 & 69.42 & 349.1 (1.02$\times$) & 51.2 & 174.0 & 62.37 & 315.8 (1.12$\times$) & 29.0 & 174.0 \\
DivPrune & 71.82 & 367.0 (0.97$\times$) & 73.3 & 175.3 & 70.61 & 354.4 (1.00$\times$) & 51.2 & 174.8 & 66.21 & 288.6 (1.23$\times$) & 29.0 & 174.7 \\
CDPruner & 72.71 & 426.3 (0.83$\times$) & 73.3 & 175.0 & 71.39 & 372.7 (0.95$\times$) & 51.2 & 174.8 & 69.13 & 303.7 (1.17$\times$) & 29.0 & 174.7 \\
VisionZip & 72.16 & 411.2 (0.86$\times$) & 73.3 & 178.0  & 71.50 & 420.8 (0.84$\times$) & 51.2 & 175.4 & 67.47 & 326.0 (1.09$\times$) & 29.0 & 174.1 \\
\textbf{\me} & \textbf{72.96} & \textbf{335.5 (1.06$\times$)}  & 73.3 & \textbf{167.0}& \textbf{73.02} & \textbf{313.3 (1.13$\times$)} & 51.2 & \textbf{173.8} & \textbf{69.38} & \textbf{274.3 (1.29$\times$)} & 29.0 & \textbf{171.6} \\
\bottomrule
\end{tabular}%
}

\label{tab:qwen3-efficiency}
\end{table*}

\section{Conclusion}

  We presented \me, a training-free visual token router that treats multimodal token pruning as
  task-conditioned evidence search. Motivated by fruit-fly optimization, \me uses text-derived odor
  cues, sparse sensing, visual lock-on, and rescue jumps to select compact visual subsets before LLM
  prefill, without modifying the vision encoder, language model, prompts, or decoding pipeline. Across
  Qwen3-VL models from 2B to 235B, \me consistently outperforms FastV, DivPrune, CDPruner, and
  VisionZip while preserving compression-aware scaling. At 20\% visual-token retention, it retains
  93.86\% of full-token performance on average, and under the fixed-fidelity view it needs only
  39.9\% visual tokens to recover 97\% of full-token performance. Additional results on Qwen2.5-VL and
  InternVL3.5 further show that the proposed search strategy transfers beyond a single backbone
  family. These findings suggest that visual token pruning should be studied not only as fixed-ratio
  compression, but also as a scaling problem: as MLLMs grow, effective pruning must preserve the
  task-relevant evidence that lets larger models realize their capacity under tight token budgets.




\bibliographystyle{plainnat}
\bibliography{reference}

@article{bai2025qwen3,
  title={Qwen3-vl technical report},
  author={Bai, Shuai and Cai, Yuxuan and Chen, Ruizhe and Chen, Keqin and Chen, Xionghui and Cheng, Zesen and Deng, Lianghao and Ding, Wei and Gao, Chang and Ge, Chunjiang and others},
  journal={arXiv preprint arXiv:2511.21631},
  year={2025}
}

@inproceedings{chen2024internvl,
  title={Internvl: Scaling up vision foundation models and aligning for generic visual-linguistic tasks},
  author={Chen, Zhe and Wu, Jiannan and Wang, Wenhai and Su, Weijie and Chen, Guo and Xing, Sen and Zhong, Muyan and Zhang, Qinglong and Zhu, Xizhou and Lu, Lewei and others},
  booktitle={Proceedings of the IEEE/CVF conference on computer vision and pattern recognition},
  pages={24185--24198},
  year={2024}
}

@misc{bai2025qwen25vltechnicalreport,
      title={Qwen2.5-VL Technical Report}, 
      author={Shuai Bai and Keqin Chen and Xuejing Liu and Jialin Wang and Wenbin Ge and Sibo Song and Kai Dang and Peng Wang and Shijie Wang and Jun Tang and Humen Zhong and Yuanzhi Zhu and Mingkun Yang and Zhaohai Li and Jianqiang Wan and Pengfei Wang and Wei Ding and Zheren Fu and Yiheng Xu and Jiabo Ye and Xi Zhang and Tianbao Xie and Zesen Cheng and Hang Zhang and Zhibo Yang and Haiyang Xu and Junyang Lin},
      year={2025},
      eprint={2502.13923},
      archivePrefix={arXiv},
      primaryClass={cs.CV},
      url={https://arxiv.org/abs/2502.13923}, 
}

@misc{chen2024fastv,
  title        = {An Image is Worth 1/2 Tokens After Layer 2: Plug-and-Play Inference Acceleration for Large Vision-Language Models},
  author       = {Chen, Liang and Zhao, Haozhe and Liu, Tianyu and Bai, Shuai and Lin, Junyang and Zhou, Chang and Chang, Baobao},
  year         = {2024},
  eprint       = {2403.06764},
  archivePrefix = {arXiv},
  primaryClass = {cs.CV},
  url          = {https://arxiv.org/abs/2403.06764}
}

@misc{yang2024visionzip,
  title        = {VisionZip: Longer is Better but Not Necessary in Vision Language Models},
  author       = {Yang, Senqiao and Chen, Yukang and Tian, Zhuotao and Wang, Chengyao and Li, Jingyao and Yu, Bei and Jia, Jiaya},
  year         = {2024},
  eprint       = {2412.04467},
  archivePrefix = {arXiv},
  primaryClass = {cs.CV},
  url          = {https://arxiv.org/abs/2412.04467}
}

@misc{alvar2025divprune,
  title        = {DivPrune: Diversity-based Visual Token Pruning for Large Multimodal Models},
  author       = {Alvar, Saeed Ranjbar and Singh, Gursimran and Akbari, Mohammad and Zhang, Yong},
  year         = {2025},
  eprint       = {2503.02175},
  archivePrefix = {arXiv},
  primaryClass = {cs.CV},
  url          = {https://arxiv.org/abs/2503.02175}
}

@misc{zhang2025cdpruner,
  title        = {Beyond Attention or Similarity: Maximizing Conditional Diversity for Token Pruning in MLLMs},
  author       = {Zhang, Qizhe and Liu, Mengzhen and Li, Lichen and Lu, Ming and Zhang, Yuan and Pan, Junwen and She, Qi and Zhang, Shanghang},
  year         = {2025},
  eprint       = {2506.10967},
  archivePrefix = {arXiv},
  primaryClass = {cs.CV},
  url          = {https://arxiv.org/abs/2506.10967}
}

@inproceedings{Guo2023EvoPromptCL,
  title={EvoPrompt: Connecting LLMs with Evolutionary Algorithms Yields Powerful Prompt Optimizers},
  author={Qingyan Guo and Rui Wang and Junliang Guo and Bei Li and Kaitao Song and Xu Tan and Guoqing Liu and Jiang Bian and Yujiu Yang and Tsinghua University and Microsoft Research},
  year={2023},
  url={https://api.semanticscholar.org/CorpusID:262012566}
}

@inproceedings{Fernando2023PromptbreederSS,
  title={Promptbreeder: Self-Referential Self-Improvement Via Prompt Evolution},
  author={Chrisantha Fernando and Dylan Banarse and Henryk Michalewski and Simon Osindero and Tim Rockt{\"a}schel},
  booktitle={International Conference on Machine Learning},
  year={2023},
  url={https://api.semanticscholar.org/CorpusID:263310323}
}

@article{Feng2024ModelSC,
  title={Model Swarms: Collaborative Search to Adapt LLM Experts via Swarm Intelligence},
  author={Shangbin Feng and Zifeng Wang and Yike Wang and Sayna Ebrahimi and Hamid Palangi and Lesly Miculicich and Achin Kulshrestha and Nathalie Rauschmayr and Yejin Choi and Yulia Tsvetkov and Chen-Yu Lee and Tomas Pfister},
  journal={ArXiv},
  year={2024},
  volume={abs/2410.11163},
  url={https://api.semanticscholar.org/CorpusID:273350735}
}

@misc{zhang2025genome,
  title        = {Nature-Inspired Population-Based Evolution of Large Language Models},
  author       = {Zhang, Yiqun and Ye, Peng and Yang, Xiaocui and Feng, Shi and Zhang, Shufei and Bai, Lei and Ouyang, Wanli and Hu, Shuyue},
  year         = {2025},
  eprint       = {2503.01155},
  archivePrefix = {arXiv},
  primaryClass = {cs.CL},
  url          = {https://arxiv.org/abs/2503.01155}
}

@inproceedings{guan2024hallusionbench,
  title     = {HallusionBench: An Advanced Diagnostic Suite for Entangled Language Hallucination and Visual Illusion in Large Vision-Language Models},
  author    = {Guan, Tianrui and Liu, Fuxiao and Wu, Xiyang and Xian, Ruiqi and Li, Zongxia and Liu, Xiaoyu and Wang, Xijun and Chen, Lichang and Huang, Furong and Yacoob, Yaser and Manocha, Dinesh and Zhou, Tianyi},
  booktitle = {Proceedings of the IEEE/CVF Conference on Computer Vision and Pattern Recognition},
  pages     = {14375--14385},
  month     = {June},
  year      = {2024}
}

@misc{fu2023mme,
  title         = {{MME}: A Comprehensive Evaluation Benchmark for Multimodal Large Language Models},
  author        = {Fu, Chaoyou and Chen, Peixian and Shen, Yunhang and Qin, Yulei and Zhang, Mengdan and Lin, Xu and Yang, Jinrui and Zheng, Xiawu and Li, Ke and Sun, Xing and Wu, Yunsheng and Ji, Rongrong and Shan, Caifeng and He, Ran},
  year          = {2023},
  eprint        = {2306.13394},
  archivePrefix = {arXiv},
  primaryClass  = {cs.CV},
  note          = {NeurIPS 2025 Datasets and Benchmarks Track Spotlight},
  url           = {https://arxiv.org/abs/2306.13394}
}

@inproceedings{kembhavi2016ai2d,
  title     = {A Diagram Is Worth a Dozen Images},
  author    = {Kembhavi, Aniruddha and Salvato, Mike and Kolve, Eric and Seo, Minjoon and Hajishirzi, Hannaneh and Farhadi, Ali},
  booktitle = {Computer Vision -- ECCV 2016},
  pages     = {235--251},
  year      = {2016},
  publisher = {Springer},
  doi       = {10.1007/978-3-319-46493-0_15},
  url       = {https://doi.org/10.1007/978-3-319-46493-0_15}
}

@misc{xai2024realworldqa,
  title        = {{Grok-1.5 Vision Preview}: {RealWorldQA} Dataset},
  author       = {{xAI}},
  year         = {2024},
  howpublished = {\url{https://x.ai/news/grok-1.5v}},
  note         = {Dataset available at \url{https://huggingface.co/datasets/xai-org/RealworldQA}; accessed 2026-05-04}
}

@inproceedings{lu2022scienceqa,
  title     = {Learn to Explain: Multimodal Reasoning via Thought Chains for Science Question Answering},
  author    = {Lu, Pan and Mishra, Swaroop and Xia, Tony and Qiu, Liang and Chang, Kai-Wei and Zhu, Song-Chun and Tafjord, Oyvind and Clark, Peter and Kalyan, Ashwin},
  booktitle = {Advances in Neural Information Processing Systems},
  volume    = {35},
  pages     = {2507--2521},
  year      = {2022},
  url       = {https://proceedings.neurips.cc/paper_files/paper/2022/hash/11332b6b6cf4485b84afadb1352d3a9a-Abstract-Conference.html}
}

@inproceedings{li2023pope,
  title     = {Evaluating Object Hallucination in Large Vision-Language Models},
  author    = {Li, Yifan and Du, Yifan and Zhou, Kun and Wang, Jinpeng and Zhao, Xin and Wen, Ji-Rong},
  booktitle = {Proceedings of the 2023 Conference on Empirical Methods in Natural Language Processing},
  pages     = {292--305},
  year      = {2023},
  address   = {Singapore},
  publisher = {Association for Computational Linguistics},
  doi       = {10.18653/v1/2023.emnlp-main.20},
  url       = {https://aclanthology.org/2023.emnlp-main.20/}
}

@inproceedings{liu2024mmbench,
  title     = {{MMBench}: Is Your Multi-modal Model an All-around Player?},
  author    = {Liu, Yuan and Duan, Haodong and Zhang, Yuanhan and Li, Bo and Zhang, Songyang and Zhao, Wangbo and Yuan, Yike and Wang, Jiaqi and He, Conghui and Liu, Ziwei and Chen, Kai and Lin, Dahua},
  booktitle = {European Conference on Computer Vision},
  pages     = {216--233},
  year      = {2024},
  publisher = {Springer},
  url       = {https://arxiv.org/abs/2307.06281}
}

@article{liu2023vsr,
  title     = {Visual Spatial Reasoning},
  author    = {Liu, Fangyu and Emerson, Guy and Collier, Nigel},
  journal   = {Transactions of the Association for Computational Linguistics},
  volume    = {11},
  pages     = {635--651},
  year      = {2023},
  publisher = {MIT Press},
  doi       = {10.1162/tacl_a_00566},
  url       = {https://aclanthology.org/2023.tacl-1.37/}
}

@inproceedings{zhu2016visual7w,
  title     = {{Visual7W}: Grounded Question Answering in Images},
  author    = {Zhu, Yuke and Groth, Oliver and Bernstein, Michael and Fei-Fei, Li},
  booktitle = {Proceedings of the IEEE Conference on Computer Vision and Pattern Recognition},
  pages     = {4995--5004},
  year      = {2016},
  doi       = {10.1109/CVPR.2016.540},
  url       = {https://ai.stanford.edu/~yukez/bibtex/zhu2016cvpr.bib}
}

@article{li2025todre,
  title = {ToDRE: Effective Visual Token Pruning via Token Diversity and Task Relevance},
  author = {Li, Duo and Yang, Zuhao and Zhang, Xiaoqin and Shao, Ling and Lu, Shijian},
  journal = {arXiv preprint arXiv:2505.18757},
  year = {2025},
  url = {https://arxiv.org/abs/2505.18757},
}

@article{Zhang2025TrimTokenatorTA,
  title={TrimTokenator: Towards Adaptive Visual Token Pruning for Large Multimodal Models},
  author={Hao Zhang and Mengsi Lyu and Chen He and Yulong Ao and Yonghua Lin},
  journal={ArXiv},
  year={2025},
  volume={abs/2509.00320},
  url={https://api.semanticscholar.org/CorpusID:281081108}
}

@article{Kim2025TrainingFreeTP,
  title={Training-Free Token Pruning via Zeroth-Order Gradient Estimation in Vision-Language Models},
  author={Youngeun Kim and Youjia Zhang and Huiling Liu and Aecheon Jung and Sunwoo Lee and Sungeun Hong},
  journal={ArXiv},
  year={2025},
  volume={abs/2509.24837},
  url={https://api.semanticscholar.org/CorpusID:281674904}
}

@article{kaplan2020scaling,
  title={Scaling Laws for Neural Language Models},
  author={Kaplan, Jared and McCandlish, Sam and Henighan, Tom and Brown, Tom B. and Chess, Benjamin and Child, Rewon and Gray, Scott and Radford, Alec and Wu, Jeffrey and Amodei, Dario},
  journal={arXiv preprint arXiv:2001.08361},
  year={2020}
}

@article{hoffmann2022chinchilla,
  title={Training Compute-Optimal Large Language Models},
  author={Hoffmann, Jordan and Borgeaud, Sebastian and Mensch, Arthur and Buchatskaya, Elena and Cai, Trevor and Rutherford, Eliza and de Las Casas, Diego and Hendricks, Lisa Anne and Welbl, Johannes and Clark, Aidan and Hennigan, Tom and Noland, Eric and Millican, Katie and van den Driessche, George and Damoc, Bogdan and Guy, Aurelia and Osindero, Simon and Simonyan, Karen and Elsen, Erich and Rae, Jack W. and Vinyals, Oriol and Sifre, Laurent},
  journal={arXiv preprint arXiv:2203.15556},
  year={2022}
}

@inproceedings{Li2024InferenceOV,
  title={Inference Optimal VLMs Need Fewer Visual Tokens and More Parameters},
  author={Kevin Y. Li and Sachin Goyal and Jo{\~a}o Dias Semedo and J. Zico Kolter},
  booktitle={International Conference on Learning Representations},
  year={2024},
  url={https://api.semanticscholar.org/CorpusID:273822030}
}

@article{li2024llavaonevision,
  title={LLaVA-OneVision: Easy Visual Task Transfer},
  author={Li, Bo and Zhang, Yuanhan and Guo, Dong and Zhang, Renrui and Li, Feng and Zhang, Hao and Zhang, Kaichen and Zhang, Peiyuan and Li, Yanwei and Liu, Ziwei and Li, Chunyuan},
  journal={arXiv preprint arXiv:2408.03326},
  year={2024}
}

@article{xu2024llavauhd,
  title={LLaVA-UHD: An LMM Perceiving Any Aspect Ratio and High-Resolution Images},
  author={Xu, Ruyi and Yao, Yuan and Guo, Zonghao and Cui, Junbo and Ni, Zanlin and Ge, Chunjiang and Chua, Tat-Seng and Liu, Zhiyuan and Sun, Maosong and Huang, Gao},
  journal={arXiv preprint arXiv:2403.11703},
  year={2024}
}

@article{Huang2026ObscureBE,
  title={Obscure but Effective: Classical Chinese Jailbreak Prompt Optimization via Bio-Inspired Search},
  author={Xunbin Huang and Simeng Qin and Xiaoshuang Jia and Ranjie Duan and Huanqian Yan and Zhitao Zeng and Fei Yang and Yang Liu and Xiaojun Jia},
  journal={ArXiv},
  year={2026},
  volume={abs/2602.22983},
  url={https://api.semanticscholar.org/CorpusID:286083604}
}

@article{Pan2012ANF,
  title={A new Fruit Fly Optimization Algorithm: Taking the financial distress model as an example},
  author={Wen-Tsao Pan},
  journal={Knowl. Based Syst.},
  year={2012},
  volume={26},
  pages={69-74},
  url={https://api.semanticscholar.org/CorpusID:12902933}
}

\medskip

\newpage
\appendix



\section{Datasets Description}
\label{app:datasets}
Table~\ref{tab:dataset} summarizes the benchmarks used in our evaluation. The suite covers hallucination and object-presence judgement, general visual question answering, scientific and diagram reasoning, multilingual multiple-choice evaluation, visual spatial reasoning, and culture-specific understanding. This diversity is important for visual token pruning because different tasks rely on different kinds of evidence: some require small localized objects or OCR-like details, while others depend on global scene context or relational information. Unless otherwise specified, we follow the official split and metric of each benchmark.

  \begin{table*}[!htbp]
    \caption{
  Detailed information of the evaluation benchmarks.
  We group the benchmarks by their primary evaluation focus, including comprehensive multimodal
  evaluation, hallucination diagnosis, science and diagram reasoning, and real-world or grounded
  visual question answering.
  All benchmarks except MME are reported with accuracy; MME follows the official scoring protocol.
  For Visual7W, we evaluate a randomly sampled subset of $5{,}000$ examples from the validation
  split.
  }
  \small
  \centering
  \resizebox{\textwidth}{!}{%
  \begin{tabular}{lllc}
  \toprule
  Benchmark & Category & Metrics & Size \\
  \midrule
  MME~\citep{fu2023mme}
  & Comprehensive perception and cognition
  & MME score
  & $2{,}374$ \\

  MMBench-en~\citep{liu2024mmbench}
  & General multimodal understanding
  & Accuracy
  & $4{,}329$ \\

  MMBench-CN~\citep{liu2024mmbench}
  & General multimodal understanding in Chinese
  & Accuracy
  & $4{,}329$ \\

  \midrule
  HallusionBench~\citep{guan2024hallusionbench}
  & Visual illusion and language hallucination
  & Accuracy
  & $951$ \\

  POPE~\citep{li2023pope}
  & Object hallucination
  & Accuracy
  & $9{,}000$ \\

  \midrule
  AI2D~\citep{kembhavi2016ai2d}
  & Diagram understanding and reasoning
  & Accuracy
  & $3{,}088$ \\

  ScienceQA-IMG~\citep{lu2022scienceqa}
  & Multimodal science question answering
  & Accuracy
  & $1{,}949$ \\

  \midrule
  RealWorldQA~\citep{xai2024realworldqa}
  & Real-world visual question answering
  & Accuracy
  & $765$ \\

  CCBench~\citep{liu2024mmbench}
  & Chinese cultural knowledge reasoning
  & Accuracy
  & $2{,}176$ \\

  VSR~\citep{liu2023vsr}
  & Visual spatial reasoning
  & Accuracy
  & $1{,}222$ \\

  Visual7W~\citep{zhu2016visual7w}
  & Grounded visual question answering
  & Accuracy
  & $5{,}000$ \\

  \bottomrule
  \end{tabular}%
  }

  \label{tab:dataset}
  \end{table*}

\section{Implementation Details and Hyperparameters}
\label{app:hyperparams}
Table~\ref{tab:f3a-hyperparams} lists the default hyperparameters used by \me. We use exactly the same values for all model families, model sizes, datasets, and retention ratios; the only quantity changed across the main compression settings is the final token budget $K=\lfloor\rho N\rceil$. The sparse sensing matrices are initialized once with the listed seed and then kept frozen. This keeps the method training-free and avoids tuning hyperparameters on individual benchmarks. We further include a hyperparameter stability analysis in Table~\ref{tab:hyperparam-stability}, showing that moderate changes to the main hyperparameter groups have limited impact on performance.

\begin{table*}[t]
\centering
\small
\setlength{\tabcolsep}{5pt}
\caption{Default hyperparameters of \me. The same setting is used throughout all main, efficiency, and ablation experiments unless explicitly stated otherwise.}
\label{tab:f3a-hyperparams}
\begin{tabular}{lll}
\toprule
Component & Hyperparameter & Value \\
\midrule
Sparse sensing heads & Number of heads $H_s$ & 16 \\
 & Shared sensing dimension $d_s$ & 128 \\
 & Non-zero entries $(n_v,n_t,n_b)$ & $(32,8,16)$ \\
 & Active heads $k_h$ & 4 \\
 & Head-gate temperature $\tau_h$ & 0.5 \\
 & Random seed & 42 \\
\midrule
Coarse search & Window size $w$ & 2 \\
 & Scaffold tokens per selected window & 1 \\
\midrule
Visual lock-on & Local neighborhood radius $r$ & 1 grid step \\
 & Spatial bandwidth $\sigma_p$ & 2 \\
 & Local-support weight $\lambda$ & 0.35 \\
 & Redundancy weight $\beta$ & 0.35 \\
\midrule
Rescue jump & Jump budget fraction $\alpha_{\mathrm{jump}}$ & 0.15 \\
 & Feature/spatial coverage balance $\alpha_c$ & 0.5 \\
 & Uncertainty weight $\gamma$ & 0.25 \\
 & Coverage penalty $\eta$ & 0.50 \\
\bottomrule
\end{tabular}
\end{table*}

We conduct a small hyperparameter stability analysis on Qwen3-VL-8B using HallusionBench, RealWorldQA, and AI2D at 40\% visual-token retention. We vary one hyperparameter group at a time while keeping all other settings fixed to Table~\ref{tab:f3a-hyperparams}. Table~\ref{tab:hyperparam-stability} reports the average accuracy over the three benchmarks and the change relative to the default setting. The performance remains stable across sensing-head count, coarse window size, rescue-jump budget, and random seed, with all tested variants staying within 0.5 accuracy points of the default configuration.

\begin{table}[t]
\centering
\small
\setlength{\tabcolsep}{5pt}
\caption{Hyperparameter stability on Qwen3-VL-8B at 40\% retention. Avg. is computed over HallusionBench, RealWorldQA, and AI2D.}
\label{tab:hyperparam-stability}
\begin{tabular}{llcc}
\toprule
Group & Setting & Avg. Acc. & $\Delta$ \\
\midrule
Default & Table~\ref{tab:f3a-hyperparams} & \textbf{69.72} & 0.00 \\
\midrule
Sensing heads & $H_s=8$ & 69.31 & -0.41 \\
 & $H_s=32$ & 69.58 & -0.14 \\
\midrule
Coarse window & $w=1$ & 69.35 & -0.37 \\
 & $w=3$ & 69.41 & -0.31 \\
\midrule
Rescue budget & $\alpha_{\mathrm{jump}}=0.10$ & 69.46 & -0.26 \\
 & $\alpha_{\mathrm{jump}}=0.20$ & 69.50 & -0.22 \\
\midrule
Random seed & seed $=7$ & 69.51 & -0.21 \\
 & seed $=123$ & 69.60 & -0.12 \\
\bottomrule
\end{tabular}
\end{table}

\section{Supplementary Experiments}
This section provides additional analyses that complement the main results. We first test whether the fixed-fidelity token-demand conclusion is sensitive to the chosen fidelity threshold. We then include the normalized ablation figure used to visualize how each component contributes under different pruning budgets. Finally, we provide the complete main-result tables for model scales: that are omitted from the main paper due to space limits.

\subsection{Fixed-Fidelity Sensitivity}
\label{app:fixed_fidelity_sensitivity}

In the main paper, we use $\tau=0.97$ as the primary near-full fidelity target.
  Here we report the same fixed-fidelity token demand under two additional targets,
  $\tau=0.95$ and $\tau=0.98$, with results summarized in
  Table~\ref{tab:fixed-fidelity-sensitivity}. For each model and method, $r_\tau$
  is estimated by linearly interpolating the measured 20\%, 40\%, 60\%, and 100\%
  retention points, where the 100\% point corresponds to full-token inference.
  Lower values indicate that fewer visual tokens are required to reach the target
  fraction of full-token performance.

\begin{table*}[t]
\centering
\small
\setlength{\tabcolsep}{5.5pt}
\caption{Sensitivity of fixed-fidelity token demand on Qwen3-VL. We report $r_{0.95}$ and $r_{0.98}$, the minimum visual token retention required to preserve 95\% and 98\% of full-token performance, respectively. Lower is better.}
\label{tab:fixed-fidelity-sensitivity}
\begin{tabular}{llrrrrr}
\toprule
Target & Model & CDPruner & FastV & DivPrune & VisionZip & \me \\
\midrule
\multirow{7}{*}{$r_{0.95}$}
& 2B & 34.3 & 43.0 & 35.9 & 37.2 & \textbf{25.7} \\
& 4B & 33.8 & 40.5 & 37.2 & 36.1 & \textbf{29.8} \\
& 8B & 32.5 & 38.8 & 39.2 & 32.1 & \textbf{26.7} \\
& 30B-A3B & 31.4 & 31.3 & 25.3 & 27.7 & \textbf{20.0} \\
& 32B & 36.1 & 51.6 & 49.1 & 43.7 & \textbf{31.4} \\
& 235B-A22B & 36.6 & 41.0 & 35.9 & 35.0 & \textbf{27.5} \\
& Avg. & 34.1 & 41.0 & 37.1 & 35.3 & \textbf{26.9} \\
\midrule
\multirow{7}{*}{$r_{0.98}$}
& 2B & 71.4 & 76.8 & 73.2 & 72.8 & \textbf{68.6} \\
& 4B & 65.2 & 70.6 & 68.9 & 67.5 & \textbf{54.5} \\
& 8B & 56.1 & 69.1 & 67.6 & 59.2 & \textbf{42.9} \\
& 30B-A3B & 58.6 & 51.5 & 47.6 & 42.1 & \textbf{42.0} \\
& 32B & 63.1 & 77.2 & 77.2 & 75.8 & \textbf{51.5} \\
& 235B-A22B & 66.2 & 73.6 & 63.1 & 61.5 & \textbf{53.1} \\
& Avg. & 63.4 & 69.8 & 66.3 & 63.1 & \textbf{52.1} \\
\bottomrule
\end{tabular}
\end{table*}

The trend is stable across both target fidelities. At the more permissive 95\% target, \me requires 26.9\% visual tokens on average, compared with 34.1\% for the strongest competing baseline. At the stricter 98\% target, \me requires 52.1\% visual tokens on average, while the strongest competing baseline requires 63.1\%. This shows that the fixed-fidelity advantage is not specific to the 97\% threshold used in the main text.

\subsection{Ablation Figure}
\label{app:ablation}
\begin{figure}[!htbp]
  \centering
  \includegraphics[width=\linewidth]{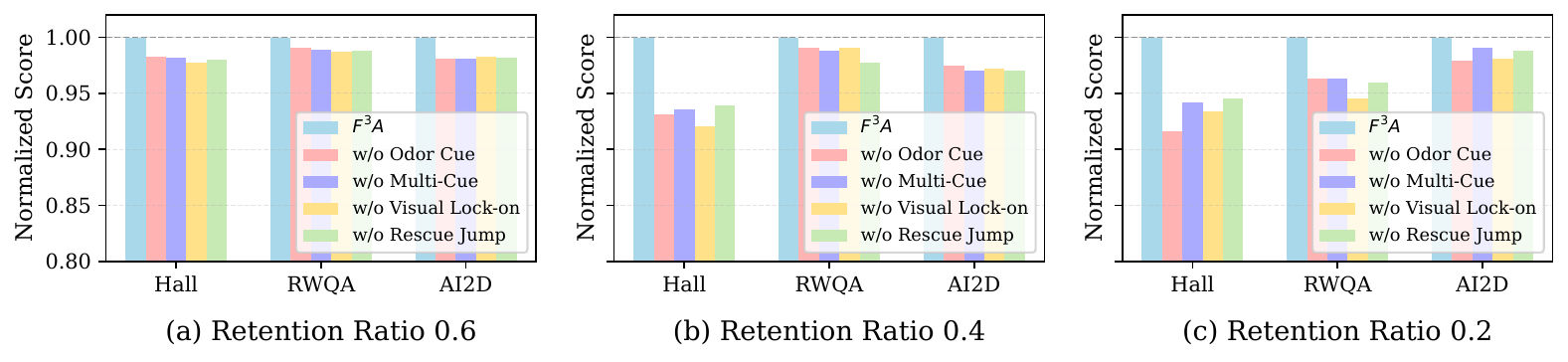}
  \caption{Normalized ablation scores on Qwen3-VL-8B over HallusionBench, RealWorldQA, and AI2D at 60\%, 40\%, and 20\% visual token retention. Scores are normalized by the full \me variant at the same retention ratio and benchmark.}
  \label{fig:main-ablation-bars}
\end{figure}

Figure~\ref{fig:main-ablation-bars} visualizes the same ablation study reported in the main text, but normalizes every variant by the full \me score at the same benchmark and retention ratio. This view makes the relative contribution of each component easier to compare across datasets with different score ranges. The drops become larger as the retention ratio decreases, indicating that the odor cue, multi-cue construction, visual lock-on, and rescue jump are most important when the token budget is tight.

\subsection{Supplementary Main Experiments}
\label{app:mainexp}
We report the full per-dataset results for model scales and backbone families not shown in the main
  paper. Tables~\ref{tab:qwen3-qwen3vl4b-results}--\ref{tab:qwen3-qwen3vl32b-results} give the
  intermediate Qwen3-VL scales, including Qwen3-VL-4B, 8B, 30B-A3B, and 32B. Together with the 2B and
  235B endpoint tables in the main paper, these results cover the full Qwen3-VL range from 2B to
  235B. Tables~\ref{tab:qwen25-7b-results} and~\ref{tab:qwen25-32b-results} report additional
  Qwen2.5-VL results, and Tables~\ref{tab:internvl35-8b-results} and~\ref{tab:internvl35-38b-results}
  report InternVL3.5 results. Across all tables, MME is reported with the official MME score, while
  the other benchmarks are reported as accuracies. The Acc. and Rel. columns follow the main-paper
  definition and exclude MME from the average.

\begin{table*}[t]
  \caption{Qwen3-VL scaling results for Qwen3-VL-4B. \textbf{Acc.}\ is the average accuracy over the non-MME datasets with a full-token baseline shown in the corresponding table, and \textbf{Rel.}\ is the ratio between this average accuracy and the corresponding full-token result.} 
  \label{tab:qwen3-qwen3vl4b-results}
  \centering
  \scriptsize
  \setlength{\tabcolsep}{1.2pt}
  \renewcommand{\arraystretch}{0.96}
  \begin{tabular*}{\textwidth}{@{\extracolsep{\fill}}llrrrrrrrrrrr|cc@{}}
  \toprule
  Ratio & Method & Hall & MME & AI2D & RWQA & SQA & POPE & MB$^{\mathrm{en}}$ & MB$^{\mathrm{zh}}$ & CCB & VSR & V7W & Acc & Rel \\
  \midrule
  100\% & Qwen3-VL-4B & 57.35 & 2274.3 & 82.06 & 71.11 & 92.71 & 87.61 & 83.99 & 83.06 & 75.37 & 82.41 & 88.62 & 80.43 & 100.00 \\
  \midrule
  \multirow{5}{*}{60\%} & CDPruner & 56.62 & 2224.4 & 77.69 & 67.45 & 88.10 & 87.38 & 83.13 & 81.84 & 74.17 & \textbf{82.90} & 86.54 & 78.58 & 97.70 \\
   & FastV & 55.98 & 2172.1 & 77.62 & 65.36 & 89.07 & 87.04 & 83.04 & \textbf{81.96} & \textbf{74.26} & 82.57 & 85.52 & 78.24 & 97.28 \\
   & DivPrune & 56.19 & 2209.5 & 77.62 & 68.89 & 88.10 & 87.42 & 82.26 & 81.47 & 73.48 & 81.75 & 86.42 & 78.36 & 97.43 \\
   & VisionZip & 54.51 & \textbf{2243.6} & 77.66 & 68.63 & 88.92 & 87.48 & 82.97 & 81.54 & 73.94 & 81.67 & \textbf{87.18} & 78.45 & 97.54 \\
   & \textbf{\me (Ours)} & \textbf{57.98} & 2232.3 & \textbf{78.63} & \textbf{69.15} & \textbf{89.79} & \textbf{87.50} & \textbf{83.64} & \textbf{81.96} & 73.85 & 82.08 & 87.14 & \textbf{79.17} & \textbf{98.44} \\
  \midrule
  \multirow{5}{*}{40\%} & CDPruner & 56.40 & 2147.7 & 76.07 & 65.62 & 85.63 & 86.77 & 81.86 & 80.68 & \textbf{73.66} & \textbf{82.41} & 85.60 & 77.47 & 96.32 \\
   & FastV & 54.83 & 2051.2 & 73.28 & 62.48 & 86.97 & 85.44 & 81.82 & 80.43 & 73.39 & 82.24 & 82.74 & 76.36 & 94.94\\
   & DivPrune & 55.67 & 2128.5 & 75.55 & \textbf{67.32} & 84.50 & 86.67 & 80.68 & 79.71 & 71.55 & 81.91 & 84.92 & 76.85 & 95.55 \\
   & VisionZip & 56.30 & 2138.8 & 75.84 & 65.49 & 84.30 & 86.96 & 82.02 & 80.38 & 73.57 & 81.10 & \textbf{85.82} & 77.18 & 95.96 \\
   & \textbf{\me (Ours)} & \textbf{58.09} & \textbf{2183.0} & \textbf{76.13} & 66.41 & \textbf{88.30} & \textbf{86.97} & \textbf{82.39} & \textbf{80.87} & 73.25 & 81.18 & 85.40 & \textbf{77.90} & \textbf{96.85} \\
  \midrule
  \multirow{5}{*}{20\%} & CDPruner & 52.20 & 1934.3 & 69.43 & 63.92 & 81.27 & 85.22 & \textbf{79.90} & 77.96 & 69.84 & 77.91 & 82.54 & 74.02 & 92.03 \\
   & FastV & 51.57 & 1832.9 & 67.29 & 52.94 & \textbf{83.68} & 79.59 & 79.21 & 77.96 & 70.40 & 77.91 & 76.90 & 71.75 & 89.20 \\
   & DivPrune & \textbf{53.99} & 1998.1 & 70.92 & 63.27 & 82.20 & 84.81 & 77.54 & 75.86 & 67.73 & 78.81 & 82.36 & 73.75 & 91.69 \\
   & VisionZip & 53.15 & 1790.9 & 67.55 & 62.88 & 83.07 & 81.80 & 78.33 & 77.57 & 69.43 & 77.09 & 81.12 & 73.20 & 91.01 \\
   & \textbf{\me (Ours)} & 53.36 & \textbf{2064.3} & \textbf{71.34} & \textbf{65.36} & 82.30 & \textbf{85.28} & 79.34 & \textbf{78.84} & \textbf{70.63} & \textbf{80.11} & \textbf{83.18} & \textbf{74.97} & \textbf{93.22} \\
  \bottomrule
  \end{tabular*}
\end{table*}

\begin{table*}[t]
  \caption{Qwen3-VL scaling results for Qwen3-VL-8B. \textbf{Acc.}\ is the average accuracy over the non-MME datasets with a full-token baseline shown in the corresponding table, and \textbf{Rel.}\ is the ratio between this average accuracy and the corresponding full-token result.} 
  \label{tab:qwen3-qwen3vl8b-results}
  \centering
  \scriptsize
  \setlength{\tabcolsep}{1.2pt}
  \renewcommand{\arraystretch}{0.96}
  \begin{tabular*}{\textwidth}{@{\extracolsep{\fill}}llrrrrrrrrrrr|cc@{}}
  \toprule
  Ratio & Method & Hall & MME & AI2D & RWQA & SQA & POPE & MB$^{\mathrm{en}}$ & MB$^{\mathrm{zh}}$ & CCB & VSR & V7W & Acc & Rel \\
  \midrule
  100\% & Qwen3-VL-8B & 61.45 & 2340.1 & 83.23 & 69.41 & 95.43 & 88.84 & 84.75 & 84.22 & 77.53 & 83.14 & 88.42 & 81.64 & 100.00 \\
  \midrule
  \multirow{5}{*}{60\%} & CDPruner & 61.56 & 2272.4 & 80.02 & 67.71 & 90.10 & 88.87 & 84.24 & \textbf{83.60} & 76.93 & 82.65 & 87.24 & 80.29 & 98.35 \\
   & FastV & 59.88 & 2247.3 & 78.98 & 65.88 & 90.51 & 88.33 & 83.16 & 82.74 & 77.49 & 83.06 & 85.28 & 79.53 & 97.41 \\
   & DivPrune & 59.88 & 2267.3 & 78.63 & 66.93 & 88.76 & 88.66 & 83.34 & 82.79 & \textbf{78.04} & 82.98 & 86.22 & 79.62 & 97.53 \\
   & VisionZip & 60.82 & \textbf{2328.4} & 79.60 & 67.06 & 90.66 & 88.59 & 83.80 & 83.36 & 76.15 & 82.98 & 87.60 & 80.06 & 98.06 \\
   & \textbf{\me (Ours)} & \textbf{62.93} & 2249.9 & \textbf{81.12} & \textbf{69.93} & \textbf{91.43} & \textbf{89.02} & \textbf{84.43} & 83.57 & 76.57 & \textbf{83.47} & \textbf{87.68} & \textbf{81.02} & \textbf{99.23} \\
  \midrule
  \multirow{5}{*}{40\%} & CDPruner & 58.62 & 2234.4 & 77.66 & 66.93 & 87.94 & \textbf{88.63} & 83.50 & 82.35 & 75.46 & 81.34 & 85.96 & 78.84 & 96.57 \\
   & FastV & 58.09 & 2138.8 & 76.30 & 64.58 & 88.61 & 85.58 & 82.19 & 82.16 & \textbf{76.84} & 82.32 & 82.12 & 77.88 & 95.39 \\
   & DivPrune & 58.83 & 2193.4 & 75.39 & 65.23 & 85.89 & 87.77 & 82.12 & 81.05 & 76.38 & 80.52 & 83.94 & 77.71 & 95.19 \\
   & VisionZip & 58.62 & 2210.0 & 75.84 & 67.71 & 88.30 & 88.18 & 82.97 & 81.98 & 75.97 & 82.24 & \textbf{86.66} & 78.85 & 96.58 \\
   & \textbf{\me (Ours)} & \textbf{62.40} & \textbf{2266.1} & \textbf{78.53} & \textbf{68.24} & \textbf{89.12} & 88.43 & \textbf{83.57} & \textbf{82.99} & 75.83 & \textbf{82.90} & 86.40 & \textbf{79.84} & \textbf{97.79} \\
  \midrule
  \multirow{5}{*}{20\%} & CDPruner & 56.72 & 2097.8 & \textbf{72.80} & 63.27 & 84.92 & \textbf{87.39} & 80.64 & \textbf{79.74} & \textbf{74.54} & 78.89 & 83.34 & 75.22 & 92.36 \\
   & FastV & 53.04 & 1863.1 & 70.43 & 55.69 & 82.71 & 78.39 & 78.05 & 77.94 & 74.17 & 77.50 & 75.78 & 72.37 & 88.64 \\
   & DivPrune & 54.30 & 1985.0 & 71.50 & 59.35 & 82.56 & 84.98 & 77.94 & 77.73 & 72.70 & 77.00 & 80.14 & 73.82 & 90.42 \\
   & VisionZip & 54.20 & 2048.6 & 70.98 & 61.18 & 84.25 & 87.03 & \textbf{81.03} & 79.58 & 73.16 & \textbf{80.61} & \textbf{83.96} & 75.60 & 92.60 \\
   & \textbf{\me (Ours)} & \textbf{57.67} & \textbf{2115.7} & 72.60 & \textbf{64.58} & \textbf{86.87} & 85.90 & 79.78 & \textbf{79.74} & 74.17 & 79.71 & 83.08 & \textbf{76.41} & \textbf{93.59} \\
  \bottomrule
  \end{tabular*}
\end{table*}

\begin{table*}[t]
  \caption{Qwen3-VL scaling results for Qwen3-VL-30B-A3B. \textbf{Acc.}\ is the average accuracy over the non-MME datasets with a full-token baseline shown in the corresponding table, and \textbf{Rel.}\ is the ratio between this average accuracy and the corresponding full-token result.} 
  \label{tab:qwen3-qwen3vl30b-results}
  \centering
  \scriptsize
  \setlength{\tabcolsep}{1.2pt}
  \renewcommand{\arraystretch}{0.96}
  \begin{tabular*}{\textwidth}{@{\extracolsep{\fill}}llrrrrrrrrrrr|cc@{}}
  \toprule
  Ratio & Method & Hall & MME & AI2D & RWQA & SQA & POPE & MB$^{\mathrm{en}}$ & MB$^{\mathrm{zh}}$ & CCB & VSR & V7W & Acc & Rel \\
  \midrule
  100\% & Qwen3-VL-30B-A3B & 61.87 & 2381.7 & 84.55 & 71.11 & 95.13 & 90.20 & 86.74 & 85.79 & 79.01 & 87.15 & 90.06 & 83.16 & 100.00 \\
  \midrule
  \multirow{5}{*}{60\%} & CDPruner & 59.35 & 2364.3 & 81.35 & 69.80 & 92.41 & 89.88 & 85.65 & 85.33 & 77.67 & 86.09 & 88.29 & 81.58 & 98.10 \\
   & FastV & \textbf{62.19} & 2311.4 & 81.51 & 70.07 & 93.18 & 89.46 & 86.07 & 85.37 & 78.45 & \textbf{87.40} & 87.90 & 82.16 & 98.80 \\
   & DivPrune & \textbf{62.19} & 2341.0 & 82.71 & 70.20 & 93.18 & 89.96 & 85.91 & 85.30 & 78.18 & 85.76 & 88.27 & 82.17 & 98.80 \\
   & VisionZip & 61.66 & 2285.6 & \textbf{82.74} & 69.93 & \textbf{93.79} & \textbf{90.31} & \textbf{86.46} & 85.56 & 78.31 & 86.82 & 88.10 & 82.37 & 99.05 \\
   & \textbf{\me (Ours)} & 61.13 & \textbf{2375.2} & 82.48 & \textbf{70.46} & 93.33 & 90.30 & 85.95 & \textbf{85.95} & \textbf{78.55} & 86.42 & \textbf{89.48} & \textbf{82.41} & \textbf{99.09} \\
  \midrule
  \multirow{5}{*}{40\%} & CDPruner & 58.61 & 2211.8 & 78.85 & 69.54 & 90.10 & 88.63 & 84.63 & 83.27 & 77.85 & 85.43 & 87.40 & 80.43 & 96.72 \\
   & FastV & 59.98 & 2235.2 & 79.63 & 68.50 & 91.48 & 87.68 & 85.10 & 84.52 & 77.95 & 85.60 & 85.48 & 80.59 & 96.91 \\
   & DivPrune & 60.50 & 2277.8 & 80.86 & 69.28 & \textbf{92.10} & 89.44 & 85.03 & 84.33 & 77.95 & 85.43 & 85.96 & 81.09 & 97.51 \\
   & VisionZip & \textbf{60.71} & 2295.8 & 80.80 & 69.54 & 92.05 & \textbf{90.04} & 85.17 & 84.06 & \textbf{78.64} & \textbf{86.33} & 86.63 & 81.40 & 97.88 \\
   & \textbf{\me (Ours)} & 60.50 & \textbf{2315.2} & \textbf{82.19} & \textbf{69.93} & 91.79 & 90.02 & \textbf{85.70} & \textbf{84.82} & 78.27 & 86.25 & \textbf{88.96} & \textbf{81.84} & \textbf{97.88} \\
  \midrule
  \multirow{5}{*}{20\%} & CDPruner & 52.83 & 2049.8 & 74.90 & 62.48 & 87.38 & 87.20 & 83.02 & 81.15 & 74.82 & 82.32 & 85.09 & 77.12 & 92.73 \\
   & FastV & 56.51 & 1997.0 & 74.32 & 63.53 & \textbf{89.79} & 83.32 & 82.51 & 82.26 & 74.95 & 82.24 & 80.12 & 76.95 & 92.54 \\
   & DivPrune & 56.40 & 2143.3 & 78.14 & 65.88 & 88.76 & 86.90 & 82.72 & 82.28 & \textbf{77.53} & 82.65 & 81.17 & 78.24 & 94.09 \\
   & VisionZip & 55.67 & 1972.6 & 74.00 & 67.06 & 87.48 & 87.97 & 82.37 & 80.66 & 75.14 & 82.41 & 82.36 & 77.51 & 93.21 \\
   & \textbf{\me (Ours)} & \textbf{57.98} & \textbf{2189.7} & \textbf{78.30} & \textbf{67.58} & 89.43 & \textbf{89.14} & \textbf{84.47} & \textbf{83.66} & 76.61 & \textbf{82.90} & \textbf{87.30} & \textbf{79.74 } & \textbf{95.88} \\
  \bottomrule
  \end{tabular*}
\end{table*}

\begin{table*}[t]
  \caption{Qwen3-VL scaling results for Qwen3-VL-32B. \textbf{Acc.}\ is the average accuracy over the non-MME datasets with a full-token baseline shown in the corresponding table, and \textbf{Rel.}\ is the ratio between this average accuracy and the corresponding full-token result.} 
  \label{tab:qwen3-qwen3vl32b-results}
  \centering
  \scriptsize
  \setlength{\tabcolsep}{1.2pt}
  \renewcommand{\arraystretch}{0.96}
  \begin{tabular*}{\textwidth}{@{\extracolsep{\fill}}llrrrrrrrrrrr|cc@{}}
  \toprule
  Ratio & Method & Hall & MME & AI2D & RWQA & SQA & POPE & MB$^{\mathrm{en}}$ & MB$^{\mathrm{zh}}$ & CCB & VSR & V7W & Acc & Rel \\
  \midrule
  100\% & Qwen3-VL-32B & 63.76 & 2501.3 & 87.63 & 75.82 & 96.72 & 89.11 & 87.64 & 86.30 & 78.13 & 88.87 & 90.00 & 84.40 & 100.00 \\
  \midrule
  \multirow{5}{*}{60\%} & CDPruner & 61.77 & 2436.9 & 84.72 & 71.11 & 93.43 & \textbf{88.96} & 86.23 & 85.26 & 77.03 & \textbf{88.46} & 88.68 & 82.57 & 97.83 \\
   & FastV & 59.87 & 2380.8 & 83.32 & 69.02 & 93.54 & 87.63 & 85.56 & 84.38 & 77.30 & 87.15 & 86.62 & 81.44 & 96.49 \\
   & DivPrune & 59.35 & 2370.9 & 81.31 & 71.63 & 90.92 & 88.79 & 84.94 & 84.22 & 77.26 & 87.89 & 88.02 & 81.43 & 96.49 \\
   & VisionZip & 59.87 & 2402.5 & 82.77 & 70.07 & 91.48 & 88.48 & 85.33 & 85.05 & 76.70 & 88.38 & 88.02 & 81.61 & 96.70 \\
   & \textbf{\me (Ours)} & \textbf{62.50} & \textbf{2494.3} & \textbf{84.84} & \textbf{74.64} & \textbf{94.97} & 88.89 & \textbf{86.46} & \textbf{85.35} & \textbf{78.22} & 88.22 & \textbf{88.86} & \textbf{83.30} & \textbf{98.69} \\
  \midrule
  \multirow{5}{*}{40\%} & CDPruner & 59.66 & 2376.8 & \textbf{83.06} & 68.76 & 90.97 & 88.31 & 84.91 & 84.20 & 76.34 & 86.50 & 87.32 & 81.00 & 95.98 \\
   & FastV & 57.66 & 2273.0 & 80.25 & 62.75 & 88.92 & 84.56 & 84.29 & 82.97 & 75.87 & 83.80 & 83.44 & 78.45 & 92.95 \\
   & DivPrune & 55.88 & 2234.3 & 77.72 & 68.10 & 86.35 & 88.19 & 83.85 & 82.37 & 76.43 & 86.25 & 86.16 & 79.13 & 93.76 \\
   & VisionZip & 56.30 & \textbf{2381.4} & 79.66 & 66.54 & 88.81 & 87.72 & \textbf{85.19} & 84.70 & 76.34 & \textbf{86.58} & 86.74 & 79.86 & 94.62 \\
   & \textbf{\me (Ours)} & \textbf{61.03} & 2355.6 & 82.55 & \textbf{73.33} & \textbf{92.10} & \textbf{88.80} & 84.98 & \textbf{84.73} & \textbf{77.30} & \textbf{86.58} & \textbf{87.82} & \textbf{81.92} & \textbf{97.07} \\
  \midrule
  \multirow{5}{*}{20\%} & CDPruner & 51.25 & 2198.1 & 73.51 & 61.96 & 86.04 & 86.09 & \textbf{83.13} & \textbf{82.53} & 74.86 & 83.72 & 84.42 & 76.75 & 90.94 \\
   & FastV & 55.77 & 2023.4 & 73.32 & 53.99 & 85.12 & 75.40 & 80.38 & 79.95 & 74.91 & 77.41 & 78.52 & 73.48 & 87.06 \\
   & DivPrune & 51.15 & 2059.5 & 72.80 & 62.35 & 81.94 & 86.13 & 79.88 & 78.84 & 73.07 & 82.41 & 82.88 & 75.14 & 89.04 \\
   & VisionZip & 54.62 & 2189.1 & 72.44 & \textbf{67.71} & 85.48 & 83.87 & 81.49 & 81.22 & 74.91 & 83.22 & 83.56 &76.85 & 91.06 \\
   & \textbf{\me (Ours)} & \textbf{56.61} & \textbf{2269.2} & \textbf{77.56} & 61.83 & \textbf{86.45} & \textbf{87.10} & 83.06 & 82.16 & \textbf{75.09} & \textbf{84.04} & \textbf{84.88} & \textbf{77.88} & \textbf{92.27} \\
  \bottomrule
  \end{tabular*}
\end{table*}

\begin{table*}[t]
  \caption{Additional backbone results for Qwen2.5-VL-7B. \textbf{Acc.}\ is the average accuracy over the non-MME datasets with a full-token baseline shown in the corresponding table, and \textbf{Rel.}\ is the ratio between this average accuracy and the corresponding full-token result.} 
  \label{tab:qwen25-7b-results}
  \centering
  \scriptsize
  \setlength{\tabcolsep}{1.2pt}
  \renewcommand{\arraystretch}{0.96}
  \begin{tabular*}{\textwidth}{@{\extracolsep{\fill}}llrrrrrrrrrrr|cc@{}}
  \toprule
  Ratio & Method & Hall & MME & AI2D & RWQA & SQA & POPE & MB$^{\mathrm{en}}$ & MB$^{\mathrm{zh}}$ & CCB & VSR & V7W & Acc & Rel \\
  \midrule
  100\% & Qwen2.5-VL-7B & 53.41 & 2251.0 & 82.25 & 67.32 & 89.12 & 87.59 & 82.10 & 82.37 & 75.01 & 81.66 & 86.72 & 78.76 & 100.00 \\
  \midrule
  \multirow{5}{*}{60\%}
  & CDPruner
  & 52.61 & 2185.7 & 78.55 & \textbf{64.63} & 84.22 & 87.15
  & 81.28 & 81.55 & \textbf{73.88} & 81.01 & 84.99
  & 76.99 & 97.77 \\
  & FastV
  & 52.77 & 2149.7 & 77.73 & 62.94 & 83.59 & 86.54
  & 80.46 & 80.72 & 73.51 & 80.84 & 83.68
  & 76.28 & 96.89 \\
  & DivPrune
  & 52.66 & 2181.2 & 78.30 & 64.49 & 83.95 & 87.22
  & 81.44 & 81.38 & 73.73 & 80.84 & 84.80
  & 76.88 & 97.64 \\
  & VisionZip
  & 53.14 & 2201.5 & 78.80 & 64.29 & 84.66 & \textbf{87.33}
  & \textbf{81.69} & 81.71 & 70.21 & 79.27 & 84.80
  & 76.59 & 97.25\\
  & \textbf{\me (Ours)}
  & \textbf{54.62} & \textbf{2213.9} & \textbf{79.57} & 62.61 & \textbf{86.81} & 86.30
  & 81.16 & \textbf{81.82} & \textbf{73.88} & \textbf{81.25} & \textbf{85.16}
  & \textbf{77.32} & \textbf{98.18} \\
  \midrule
  \multirow{5}{*}{40\%}
  & CDPruner
  & 51.54 & 2125.0 & \textbf{76.90} & \textbf{63.62} & 82.44 & 86.71
  & 80.46 & 80.31 & 73.13 & 80.03 & \textbf{84.12}
  & 75.93 & 96.40 \\
  & FastV
  & 51.17 & 2057.4 & 75.26 & 61.60 & 81.81 & 85.10
  & 79.64 & 80.06 & 72.76 & 80.03 & 82.50
  & 74.99 & 95.21 \\
  & DivPrune
  & 51.70 & 2109.2 & 76.49 & 63.28 & 82.72 & 86.89
  & 80.05 & 80.56 & 73.36 & 80.19 & 83.68
  & 75.89 & 96.36 \\
  & VisionZip
  & 52.07 & 2127.2 & \textbf{76.90} & 63.44 & 83.08 & \textbf{87.17}
  & \textbf{80.66} & 80.72 & 68.65 & 79.22 & 83.68
  & 75.56 & 95.94 \\
  & \textbf{\me (Ours)}
  & \textbf{53.35} & \textbf{2166.5} & 76.85 & 61.05 & \textbf{84.76} & 84.56
  & 80.52 & \textbf{81.94} & \textbf{73.51} & \textbf{80.70} & \textbf{84.12}
  & \textbf{76.14} & \textbf{96.67} \\
  \midrule
  \multirow{5}{*}{20\%}
  & CDPruner
  & 49.40 & \textbf{1994.4} & 71.97 & 59.91 & 78.90 & 84.50
  & 78.41 & 77.84 & 71.26 & 77.58 & 82.38
  & 73.22 & 92.90 \\
  & FastV
  & 47.27 & 1868.3 & 69.50 & 55.88 & 77.09 & 78.20
  & 76.35 & 76.60 & 70.13 & 75.13 & 79.35
  & 70.55 & 89.48 \\
  & DivPrune
  & 48.87 & 1992.1 & \textbf{72.22} & 60.25 & 79.32 & 84.52
  & 77.58 & 77.43 & 70.88 & 77.41 & 81.52
  & 73.00 & 92.61 \\
  & VisionZip
  & 48.87 & 1917.9 & 71.15 & \textbf{60.92} & 79.76 & \textbf{85.14}
  & 78.00 & 77.59 & \textbf{71.63} & 77.99 & 81.95
  & 73.30 & 92.99 \\
  & \textbf{\me (Ours)}
  & \textbf{49.57} & 1989.0 & 70.95 & 56.86 & \textbf{82.91} & 82.38
  & \textbf{81.00} & \textbf{80.24} & 69.02 & \textbf{78.48} & \textbf{82.82}
  & \textbf{73.42} & \textbf{93.03} \\
  \bottomrule
  \end{tabular*}
\end{table*}

\begin{table*}[t]
  \caption{Additional backbone results for Qwen2.5-VL-32B. \textbf{Acc.}\ is the average accuracy over the non-MME datasets with a full-token baseline shown in the corresponding table, and \textbf{Rel.}\ is the ratio between this average accuracy and the corresponding full-token result.} 
  \label{tab:qwen25-32b-results}
  \centering
  \scriptsize
  \setlength{\tabcolsep}{1.2pt}
  \renewcommand{\arraystretch}{0.96}
  \begin{tabular*}{\textwidth}{@{\extracolsep{\fill}}llrrrrrrrrrrr|cc@{}}
  \toprule
  Ratio & Method & Hall & MME & AI2D & RWQA & SQA & POPE & MB$^{\mathrm{en}}$ & MB$^{\mathrm{zh}}$ & CCB & VSR & V7W & Acc & Rel \\
  \midrule
  100\% & Qwen2.5-VL-32B & 61.72 & 2403.9 & 84.26 & 68.89 & 91.33 & 88.09 & 85.30 & 84.65 & 75.23 & 86.61 & 88.39 & 81.45 & 100.00 \\
  \midrule
  \multirow{5}{*}{60\%}
  & CDPruner
  & 59.25 & 2387.1 & 81.31 & 67.51 & 88.41 & 87.74
  & 82.28 & 83.14 & 74.03 & 85.57 & 86.62
  & 79.59 & 97.72 \\
  & FastV
  & 59.73 & 2331.8 & 81.06 & 67.86 & 89.05 & 87.43
  & 83.62 & 83.23 & \textbf{74.33} & \textbf{86.87} & 86.27
  & 79.94 & 98.16 \\
  & DivPrune
  & 58.77 & 2363.0 & 82.41 & \textbf{68.00} & 89.05 & 87.83
  & 83.38 & 83.14 & 74.10 & 85.26 & 86.62
  & 79.86 & 97.98 \\
  & VisionZip
  & \textbf{61.53} & 2305.3 & \textbf{82.49} & 67.72 & 89.12 & 87.39
  & 83.04 & 82.40 & 74.18 & 86.26 & 86.62
  & 80.08 & 98.32 \\
  & \textbf{\me (Ours)}
  & 61.21 & \textbf{2396.7} & 81.51 & 63.92 & \textbf{89.50} & \textbf{88.18}
  & \textbf{84.70} & \textbf{84.36} & 74.28 & 86.60 & \textbf{87.26}
  & \textbf{80.15} & \textbf{98.41} \\
  \midrule
  \multirow{5}{*}{40\%}
  & CDPruner
  & 58.51 & 2233.2 & 78.78 & \textbf{67.38} & 86.58 & 86.70
  & 81.25 & 82.28 & 73.57 & 84.88 & 85.74
  & 78.57 & 96.46 \\
  & FastV
  & 59.87 & 2254.9 & 79.37 & 66.48 & 87.86 & 85.65
  & 81.76 & 83.38 & 73.65 & 85.05 & 83.97
  & 78.70 & 96.63 \\
  & DivPrune
  & 60.36 & 2298.1 & \textbf{80.64} & 67.17 & 88.14 & 87.43
  & 83.59 & 83.34 & 73.65 & 84.88 & 84.50
  & 79.37 & 97.45 \\
  & VisionZip
  & 58.77 & 2217.4 & 77.88 & 62.61 & 87.12 & 86.96
  & 82.93 & 81.79 & 71.13 & 84.45 & \textbf{86.14}
  & 77.98 & 95.74 \\
  & \textbf{\me (Ours)}
  & \textbf{60.55} & \textbf{2314.9} & \textbf{80.64} & \textbf{67.38} & \textbf{88.41} & \textbf{87.92}
  & \textbf{83.76} & \textbf{83.03} & \textbf{74.10} & \textbf{85.83} & 85.30
  & \textbf{79.69} & \textbf{97.85} \\
  \midrule
  \multirow{5}{*}{20\%}
  & CDPruner
  & 52.04 & 2067.4 & 72.09 & 55.29 & 84.40 & 84.32
  & 79.90 & 80.51 & 68.42 & 81.73 & 83.42
  & 74.21 & 91.12 \\
  & FastV
  & \textbf{56.47} & 2014.5 & 74.15 & 61.66 & 86.22 & 73.72
  & 81.21 & 81.09 & 71.47 & 81.85 & 79.73
  & 74.76 & 91.79 \\
  & DivPrune
  & 56.35 & 2139.5 & \textbf{77.94} & 63.93 & 85.12 & 85.04
  & 81.46 & 81.09 & 70.35 & 82.28 & 80.88
  & 76.44 & 93.86 \\
  & VisionZip
  & 55.86 & 1990.4 & 73.98 & 60.10 & 83.84 & \textbf{85.91}
  & 81.21 & 79.99 & 71.24 & \textbf{82.71} & 83.53
  & 75.84 & 93.11 \\
  & \textbf{\me (Ours)}
  & 52.96 & \textbf{2190.0} & 74.82 & \textbf{65.76} & \textbf{86.84} & 85.48
  & \textbf{81.72} & \textbf{81.59} & \textbf{71.32} & 82.11 & \textbf{84.15}
  & \textbf{76.67} & \textbf{94.14} \\
  \bottomrule
  \end{tabular*}
\end{table*}

\begin{table*}[t]
  \caption{Additional backbone results for InternVL3.5-8B. \textbf{Acc.}\ is the average accuracy over the non-MME datasets with a full-token baseline shown in the corresponding table, and \textbf{Rel.}\ is the ratio between this average accuracy and the corresponding full-token result.} 
  \label{tab:internvl35-8b-results}
  \centering
  \scriptsize
  \setlength{\tabcolsep}{1.2pt}
  \renewcommand{\arraystretch}{0.96}
  \begin{tabular*}{\textwidth}{@{\extracolsep{\fill}}llrrrrrrrrrrr|cc@{}}
  \toprule
  Ratio & Method & Hall & MME & AI2D & RWQA & SQA & POPE & MB$^{\mathrm{en}}$ & MB$^{\mathrm{zh}}$ & CCB & VSR & V7W & Acc & Rel \\
  \midrule
  100\% & InternVL3.5-8B & 54.77 & 2371.9 & 82.67 & 65.62 & 89.00 & 88.54 & 83.61 & 83.07 & 76.22 & 79.30 & 87.74 & 79.05 & 100.00 \\
  \midrule
  \multirow{5}{*}{60\%}
  & CDPruner & 52.35 & 2276.1 & 76.33 & 63.34 & \textbf{85.07} & 87.00 & 82.17 & 82.09 & 75.74 & 77.66 & 86.06 & 76.78 & 97.12 \\
  & FastV & 53.35 & 2264.9 & 78.45 & 62.27 & 84.37 & 87.01 & 82.02 & 81.57 & 74.14 & 78.21 & 84.67 & 76.61 & 96.90 \\
  & DivPrune & 53.35 & \textbf{2298.3} & 78.12 & 63.26 & 82.77 & 86.54 & 82.19 & 81.66 & 74.75 & 78.14 & 85.55 & 76.63 & 96.94 \\
  & VisionZip & 54.22 & 2260.0 & 79.03 & 63.39 & 84.55 & 86.28 & 82.69 & 82.24 & 74.85 & \textbf{78.98} & \textbf{86.62} & 77.28 & 97.76 \\
  & \textbf{\me (Ours)} & \textbf{54.88} & 2291.7 & \textbf{79.45} & \textbf{63.98} & 84.02 & \textbf{87.10} & \textbf{83.11} & \textbf{82.49} & \textbf{75.61} & 78.82 & \textbf{86.62} & \textbf{77.61} & \textbf{98.17} \\
  \midrule
  \multirow{5}{*}{40\%}
  & CDPruner & 51.83 & 2164.9 & 74.19 & 62.88 & 81.99 & \textbf{87.68} & 81.34 & 81.01 & 74.28 & 75.61 & 84.24 & 75.51 & 95.51 \\
  & FastV & 51.76 & \textbf{2167.9} & 75.81 & 61.03 & \textbf{82.59} & 85.27 & 81.10 & 81.08 & 73.53 & 78.51 & 81.51 & 75.22 & 95.15 \\
  & DivPrune & \textbf{52.42} & 2122.4 & 74.90 & 61.68 & 80.10 & 86.83 & 81.10 & 80.58 & 74.08 & 77.89 & 83.27 & 75.28 & 95.23 \\
  & VisionZip & 52.25 & 2131.4 & 75.31 & 61.98 & 82.33 & 86.92 & 81.85 & 80.83 & 74.15 & \textbf{78.67} & \textbf{85.98} & 76.03 & 96.17 \\
  & \textbf{\me (Ours)} & 52.25 & 2135.7 & \textbf{77.13} & \textbf{62.26} & 82.06 & 87.54 & \textbf{82.36} & \textbf{81.82} & \textbf{74.70}  & 77.71 & 85.11 & \textbf{76.29} & \textbf{96.51} \\
  \midrule
  \multirow{5}{*}{20\%}
  & CDPruner & 48.46 & 2101.5 & 69.27 & 55.42 & 79.30 & 86.50 & 77.98 & 78.06 & 72.77 & 72.59 & 80.74 & 72.11 & 91.21 \\
  & FastV & 47.27 & 1968.7 & 69.94 & 52.63 & 76.99 & 78.09 & 77.00 & 76.84 & 72.96 & 72.96 & 77.21 & 70.19 & 88.79 \\
  & DivPrune & 48.42 & 2099.1 & 71.01 & 59.39 & 77.61 & 84.74 & 78.18 & 77.67 & 71.49 & 74.54 & 80.28 & 72.33 & 91.50 \\
  & VisionZip & 48.31 & 2051.7 & 70.52 & 58.07 & 78.77 & 86.77 & 78.85 & 78.50 & 72.03 & \textbf{76.29} & 80.48 & 72.86 & 92.16 \\
  & \textbf{\me (Ours)} & \textbf{50.55} & \textbf{2115.7} & \textbf{72.34} & \textbf{59.85} & \textbf{80.21} & \textbf{87.12} & \textbf{79.60} & \textbf{79.50} & \textbf{73.27} & 75.34 & \textbf{81.95} & \textbf{73.97} & \textbf{93.57} \\
  \bottomrule
  \end{tabular*}
\end{table*}

\begin{table*}[t]
  \caption{Additional backbone results for InternVL3.5-38B. \textbf{Acc.}\ is the average accuracy over the non-MME datasets with a full-token baseline shown in the corresponding table, and \textbf{Rel.}\ is the ratio between this average accuracy and the corresponding full-token result.} 
  \label{tab:internvl35-38b-results}
  \centering
  \scriptsize
  \setlength{\tabcolsep}{1.2pt}
  \renewcommand{\arraystretch}{0.96}
  \begin{tabular*}{\textwidth}{@{\extracolsep{\fill}}llrrrrrrrrrrr|cc@{}}
  \toprule
  Ratio & Method & Hall & MME & AI2D & RWQA & SQA & POPE & MB$^{\mathrm{en}}$ & MB$^{\mathrm{zh}}$ & CCB & VSR & V7W & Acc & Rel \\
  \midrule
  100\% & InternVL3.5-38B & 59.08 & 2489.7 & 87.05 & 73.99 & 96.20 & 90.01 & 85.47 & 84.96 & 77.96 & 84.04 & 90.56 & 82.93 & 100.00 \\
  \midrule
  \multirow{5}{*}{60\%}
  & CDPruner & 56.66 & 2417.8 & 83.74 & 72.66 & 93.41 & 88.65 & 84.36 & 84.45 & 76.64 & 83.04 & \textbf{88.75} & 81.24 & 97.91 \\
  & FastV & 58.38 & 2426.6 & 83.92 & 72.88 & 93.80 & 88.29 & 84.79 & 83.54 & 77.03 & 83.32 & 87.39 & 81.33 & 98.07 \\
  & DivPrune & 58.71 & 2389.7 & 85.13 & \textbf{73.05} & 93.80 & 88.74 & 84.55 & 84.45 & 76.79 & 82.72 & \textbf{88.75} & 81.67 & 98.52 \\
  & VisionZip & 58.35 & 2403.6 & 84.20 & 69.28 & \textbf{95.07} & 88.03 & 85.01 & 84.17 & 76.96 & 83.22 & 88.36 & 81.27 & 97.96 \\
  & \textbf{\me (Ours)} & \textbf{58.90} & \textbf{2430.9} & \textbf{85.22} & 72.74 & 94.85 & \textbf{89.10} & \textbf{85.21} & \textbf{84.71} & \textbf{77.27} & \textbf{83.70} & 88.57 & \textbf{82.03} & \textbf{98.94} \\
  \midrule
  \multirow{5}{*}{40\%}
  & CDPruner & 55.95 & 2321.4 & 81.22 & 69.36 & 91.10 & 88.48 & 83.42 & 82.58 & 76.24 & 82.36 & 86.84 & 79.75 & 96.12 \\
  & FastV & 57.31 & 2235.6 & 82.00 & \textbf{71.25} & 92.54 & 86.86 & 83.93 & 82.69 & 76.32 & 82.53 & 86.03 & 80.15 & 96.64 \\
  & DivPrune & 57.78 & 2318.2 & \textbf{83.22} & 70.07 & \textbf{93.12} & 87.29 & 83.76 & 83.52 & 76.32 & 82.36 & 86.58 & 80.40 & 96.97 \\
  & VisionZip & \textbf{58.77} & 2341.8 & 80.96 & 67.58 & 91.94 & 88.35 & \textbf{84.01} & 82.91 & 75.69 & 81.83 & 86.71 & 79.88 & 96.31 \\
  & \textbf{\me (Ours)} & 57.96 & \textbf{2365.3} & \textbf{83.22} & 69.36 & \textbf{93.12} & \textbf{88.83} & 83.93 & \textbf{83.35} & \textbf{77.57} & \textbf{83.28} & \textbf{87.39} & \textbf{80.80} & \textbf{97.43} \\
  \midrule
  \multirow{5}{*}{20\%}
  & CDPruner & 50.45 & 2089.6 & 77.13 & 65.04 & 88.41 & 86.31 & 81.80 & 80.88 & 73.91 & 79.42 & 82.21 & 76.56 & 92.07 \\
  & FastV & 54.06 & 2158.3 & 76.60 & 66.22 & 88.81 & 85.16 & 81.37 & 80.39 & 74.06 & 80.12 & 81.69 & 76.85 & 92.63 \\
  & DivPrune & 53.94 & 2097.1 & \textbf{80.43} & 67.77 & 88.41 & 86.76 & 81.62 & 80.03 & 73.24 & 79.84 & 82.86 & 77.49 & 93.44 \\
  & VisionZip & \textbf{55.04} & 2146.8 & 75.68 & 60.92 & 84.86 & 86.61 & 81.73 & 81.01 & 73.37 & 79.30 & 83.37 & 76.19 & 91.84 \\
  & \textbf{\me (Ours)} & 53.47 & \textbf{2230.7} & 76.43 & \textbf{68.52} & \textbf{89.70} & \textbf{86.86} & \textbf{81.92} & \textbf{81.29} & \textbf{74.45} & \textbf{80.87} & \textbf{85.58} & \textbf{77.91} & \textbf{93.94} \\
  \bottomrule
  \end{tabular*}
\end{table*}

\subsection{Significance Test}
  \label{app:significance}

  Using the averaged results from the three repeated runs, we conduct a paired significance analysis
  across 30 model--retention settings covering Qwen3-VL, Qwen2.5-VL, and InternVL3.5. For each
  setting,
  we compare the Acc. of \me with the strongest non-\me baseline under the same model and retention
  ratio. As shown in Table~\ref{tab:significance-test}, \me improves over the strongest baseline in
  all
  30 pairs, with a mean gain of 0.60 accuracy points and a median gain of 0.51 points. A two-sided
  sign
  test gives $p=1.9\times10^{-9}$, indicating that the advantage is consistent across model families,
  model scales, and compression budgets.

\begin{table*}[t]
  \centering
  \scriptsize
  \setlength{\tabcolsep}{4.0pt}
  \renewcommand{\arraystretch}{1.05}
  \caption{Paired significance analysis across Qwen3-VL, Qwen2.5-VL, and InternVL3.5. For each
  model--retention setting, we report the average Acc. of all pruning methods. The underlined value is
  the strongest non-\me baseline in the same setting, and Gain denotes the Acc. improvement of \me
  over that strongest baseline.}
  \label{tab:significance-test}
  \begin{tabular}{llrrrrrr}
  \toprule
  Model & Retention & CDPruner & FastV & DivPrune & VisionZip & \me & Gain \\
  \midrule
  Qwen3-VL-2B & 60\% & \underline{72.79} & 72.31 & 72.65 & 72.69 & \textbf{72.98} & +0.19 \\
  Qwen3-VL-2B & 40\% & \underline{71.78} & 70.94 & 71.67 & 71.69 & \textbf{72.38} & +0.60 \\
  Qwen3-VL-2B & 20\% & \underline{69.55} & 66.22 & 69.17 & 67.85 & \textbf{70.66} & +1.11 \\
  \midrule
  Qwen3-VL-4B & 60\% & \underline{78.58} & 78.24 & 78.36 & 78.45 & \textbf{79.17} & +0.59 \\
  Qwen3-VL-4B & 40\% & \underline{77.47} & 76.36 & 76.85 & 77.18 & \textbf{77.90} & +0.43 \\
  Qwen3-VL-4B & 20\% & \underline{74.02} & 71.75 & 73.75 & 73.20 & \textbf{74.97} & +0.95 \\
  \midrule
  Qwen3-VL-8B & 60\% & \underline{80.29} & 79.53 & 79.62 & 80.06 & \textbf{81.02} & +0.73 \\
  Qwen3-VL-8B & 40\% & 78.84 & 77.88 & 77.71 & \underline{78.85} & \textbf{79.84} & +0.99 \\
  Qwen3-VL-8B & 20\% & 75.22 & 72.37 & 73.82 & \underline{75.60} & \textbf{76.41} & +0.81 \\
  \midrule
  Qwen3-VL-30B-A3B & 60\% & 81.58 & 82.16 & 82.17 & \underline{82.37} & \textbf{82.41} & +0.04 \\
  Qwen3-VL-30B-A3B & 40\% & 80.43 & 80.59 & 81.09 & \underline{81.40} & \textbf{81.84} & +0.44 \\
  Qwen3-VL-30B-A3B & 20\% & 77.12 & 76.95 & \underline{78.24} & 77.51 & \textbf{79.74} & +1.50 \\
  \midrule
  Qwen3-VL-32B & 60\% & \underline{82.57} & 81.44 & 81.43 & 81.61 & \textbf{83.30} & +0.73 \\
  Qwen3-VL-32B & 40\% & \underline{81.00} & 78.45 & 79.13 & 79.86 & \textbf{81.92} & +0.92 \\
  Qwen3-VL-32B & 20\% & 76.75 & 73.48 & 75.14 & \underline{76.85} & \textbf{77.88} & +1.03 \\
  \midrule
  Qwen3-VL-235B-A22B & 60\% & 84.56 & 84.00 & 84.74 & \underline{84.81} & \textbf{85.39} & +0.58 \\
  Qwen3-VL-235B-A22B & 40\% & 82.75 & 82.20 & 82.90 & \underline{83.09} & \textbf{83.93} & +0.84 \\
  Qwen3-VL-235B-A22B & 20\% & \underline{80.01} & 77.25 & 79.87 & 79.86 & \textbf{81.38} & +1.37 \\
  \midrule
  Qwen2.5-VL-7B & 60\% & \underline{76.99} & 76.28 & 76.88 & 76.59 & \textbf{77.32} & +0.33 \\
  Qwen2.5-VL-7B & 40\% & \underline{75.93} & 74.99 & 75.89 & 75.56 & \textbf{76.14} & +0.21 \\
  Qwen2.5-VL-7B & 20\% & 73.22 & 70.55 & 73.00 & \underline{73.30} & \textbf{73.42} & +0.12 \\
  \midrule
  Qwen2.5-VL-32B & 60\% & 79.59 & 79.94 & 79.86 & \underline{80.08} & \textbf{80.15} & +0.07 \\
  Qwen2.5-VL-32B & 40\% & 78.57 & 78.70 & \underline{79.37} & 77.98 & \textbf{79.69} & +0.32 \\
  Qwen2.5-VL-32B & 20\% & 74.21 & 74.76 & \underline{76.44} & 75.84 & \textbf{76.67} & +0.23 \\
  \midrule
  InternVL3.5-8B & 60\% & 76.78 & 76.61 & 76.63 & \underline{77.28} & \textbf{77.61} & +0.33 \\
  InternVL3.5-8B & 40\% & 75.51 & 75.22 & 75.28 & \underline{76.03} & \textbf{76.29} & +0.26 \\
  InternVL3.5-8B & 20\% & 72.11 & 70.19 & 72.33 & \underline{72.86} & \textbf{73.97} & +1.11 \\
  \midrule
  InternVL3.5-38B & 60\% & 81.24 & 81.33 & \underline{81.67} & 81.27 & \textbf{82.03} & +0.36 \\
  InternVL3.5-38B & 40\% & 79.75 & 80.15 & \underline{80.40} & 79.88 & \textbf{80.80} & +0.40 \\
  InternVL3.5-38B & 20\% & 76.56 & 76.85 & \underline{77.49} & 76.19 & \textbf{77.91} & +0.42 \\
  \midrule
  \multicolumn{2}{l}{Mean Acc.} & 77.53 & 76.59 & 77.45 & 77.53 & \textbf{78.50} & +0.60 \\
  \multicolumn{2}{l}{Median Acc.} & 77.30 & 76.90 & 77.60 & 77.40 & \textbf{78.54} & +0.51 \\
  \multicolumn{2}{l}{Mean gain of \me over each baseline} & +0.98 & +1.91 & +1.05 & +0.98 & +0.00 &
  \textbf{+0.60} \\
  \multicolumn{2}{l}{Positive pairs over strongest baseline} & \multicolumn{5}{c}{30/30} &
  \textbf{30/30} \\
  \multicolumn{2}{l}{Two-sided sign test over strongest baseline} & \multicolumn{5}{c}
  {$p=1.9\times10^{-9}$} & \textbf{$p=1.9\times10^{-9}$} \\
  \bottomrule
  \end{tabular}
\end{table*}

\section{Case Study}
Table~\ref{tab:case_study} provides qualitative examples comparing the visual evidence retained by different pruning methods. These cases are selected to highlight scenarios where the answer depends on localized or easily missed evidence, such as clothing color, a specific object attribute, or the spatial relation between an object and its container. Compared with one-shot saliency or diversity-based pruning, \me more consistently preserves the region that supports the final answer, which explains why it can maintain accuracy under aggressive visual token compression.

\begin{table*}[t]
\centering
\caption{Qualitative case studies. Each row pair shows the original image alongside visual token selection heatmaps from five pruning methods, and whether each method answers correctly.}
\label{tab:case_study}
\setlength{\tabcolsep}{3pt}
\renewcommand{\arraystretch}{1.3}
\resizebox{\textwidth}{!}{
\begin{tabular}{p{0.19\linewidth} p{0.19\linewidth} p{0.19\linewidth} p{0.19\linewidth} p{0.19\linewidth} p{0.19\linewidth}}
\toprule
\multicolumn{1}{c}{\textbf{Original}} &
\multicolumn{1}{c}{\textbf{FastV}} &
\multicolumn{1}{c}{\textbf{DivPrune}} &
\multicolumn{1}{c}{\textbf{CDPruner}} &
\multicolumn{1}{c}{\textbf{VisionZip}} &
\multicolumn{1}{c}{\textbf{\me(ours)}} \\
\midrule
\includegraphics[width=\linewidth]{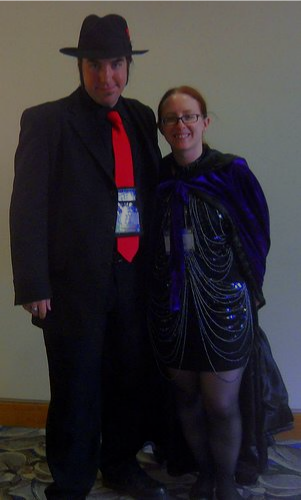} &
\includegraphics[width=\linewidth]{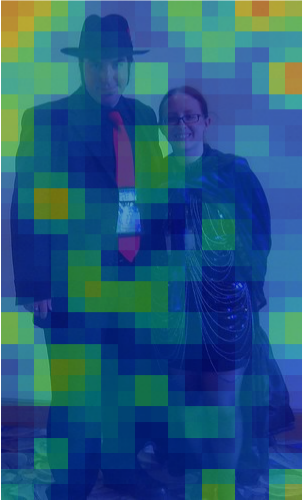} &
\includegraphics[width=\linewidth]{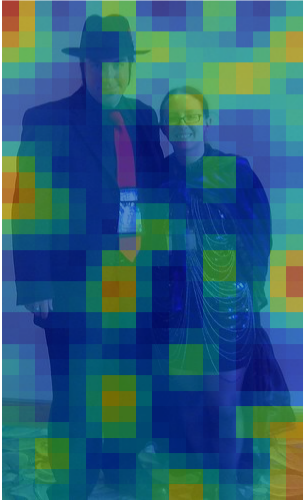} &
\includegraphics[width=\linewidth]{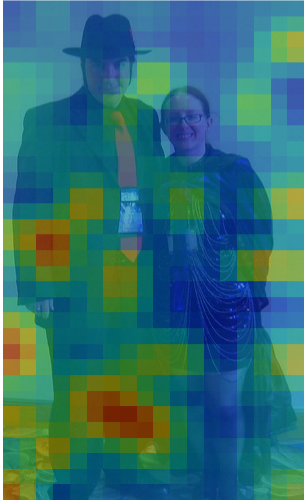} &
\includegraphics[width=\linewidth]{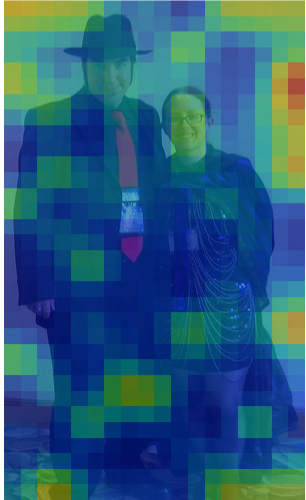} &
\includegraphics[width=\linewidth]{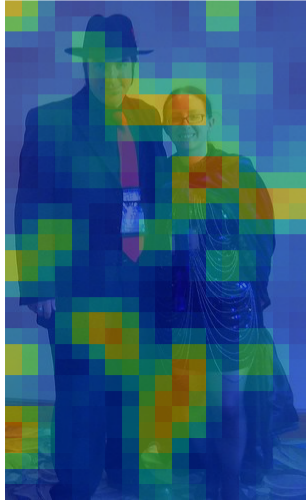} \\
\begin{minipage}[t]{\linewidth}\small
  \textbf{Q:} What color cape is the woman wearing?
\end{minipage} &
\begin{minipage}[t]{\linewidth}\centering\small Brown\\\xmark\end{minipage} &
\begin{minipage}[t]{\linewidth}\centering\small Black\\\xmark\end{minipage} &
\begin{minipage}[t]{\linewidth}\centering\small Black\\\xmark\end{minipage} &
\begin{minipage}[t]{\linewidth}\centering\small Black\\\xmark\end{minipage} &
\begin{minipage}[t]{\linewidth}\centering\small Purple\\\cmark\end{minipage} \\
\midrule
\includegraphics[width=\linewidth]{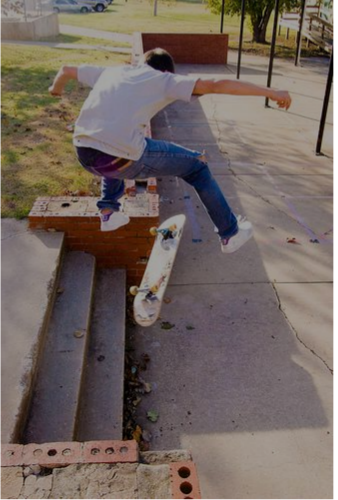} &
\includegraphics[width=\linewidth]{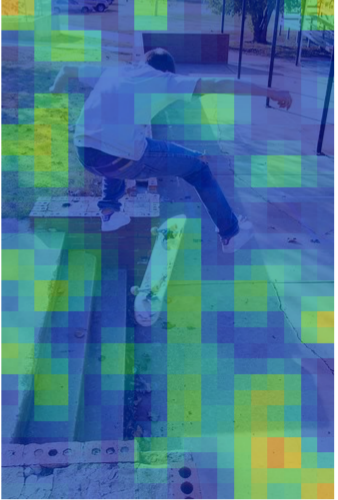} &
\includegraphics[width=\linewidth]{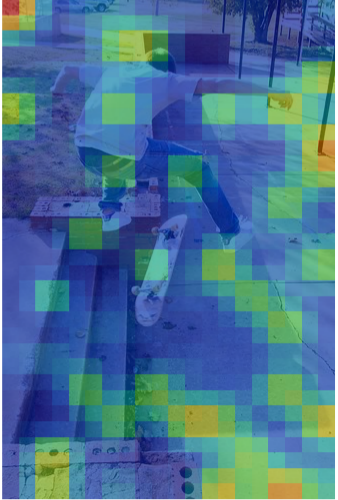} &
\includegraphics[width=\linewidth]{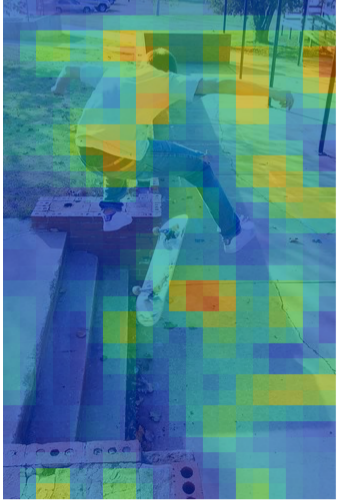} &
\includegraphics[width=\linewidth]{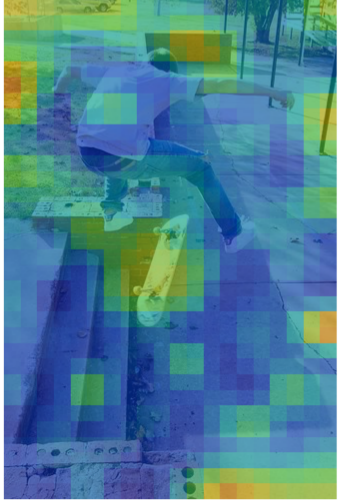} &
\includegraphics[width=\linewidth]{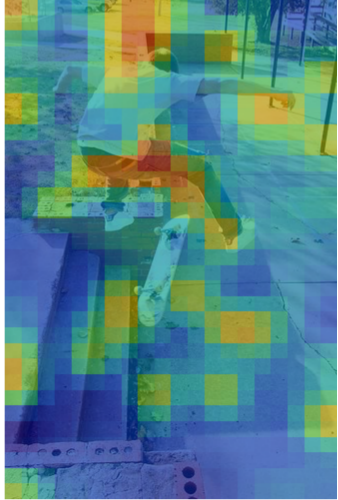} \\
\begin{minipage}[t]{\linewidth}\small
  \textbf{Q:} What is blue in the picture?
\end{minipage} &
\begin{minipage}[t]{\linewidth}\centering\small Socks\\\xmark\end{minipage} &
\begin{minipage}[t]{\linewidth}\centering\small Sneakers\\\xmark\end{minipage} &
\begin{minipage}[t]{\linewidth}\centering\small Socks\\\xmark\end{minipage} &
\begin{minipage}[t]{\linewidth}\centering\small Socks\\\xmark\end{minipage} &
\begin{minipage}[t]{\linewidth}\centering\small Pants\\\cmark\end{minipage} \\
\midrule
\includegraphics[width=\linewidth]{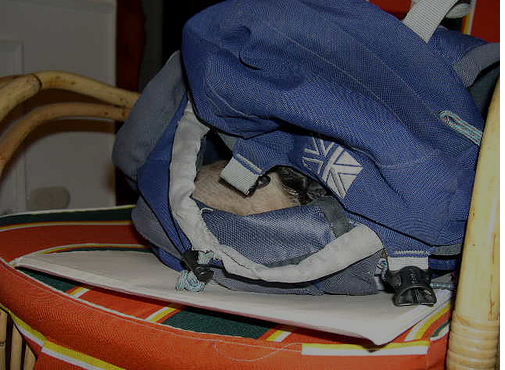} &
\includegraphics[width=\linewidth]{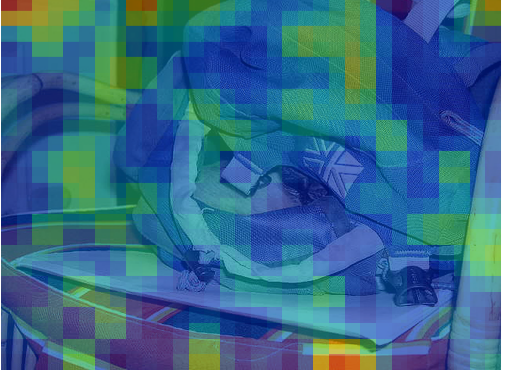} &
\includegraphics[width=\linewidth]{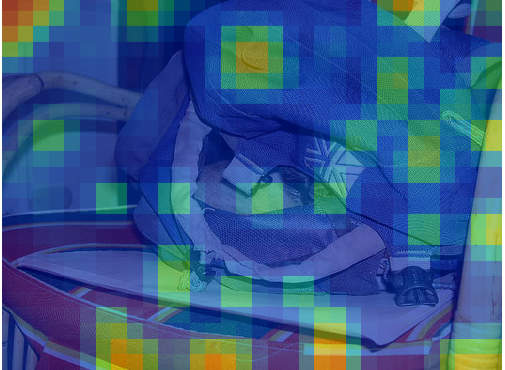} &
\includegraphics[width=\linewidth]{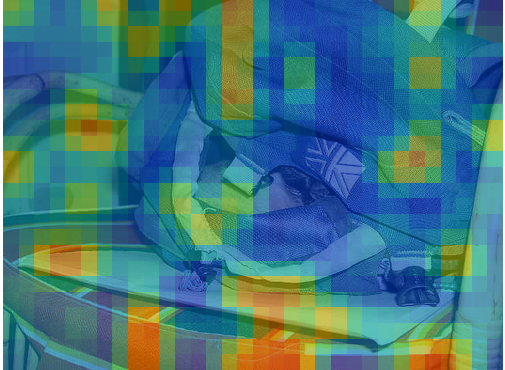} &
\includegraphics[width=\linewidth]{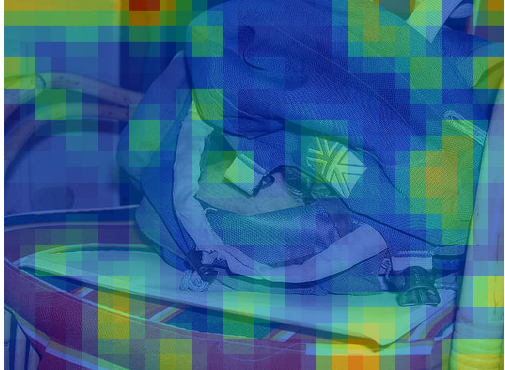} &
\includegraphics[width=\linewidth]{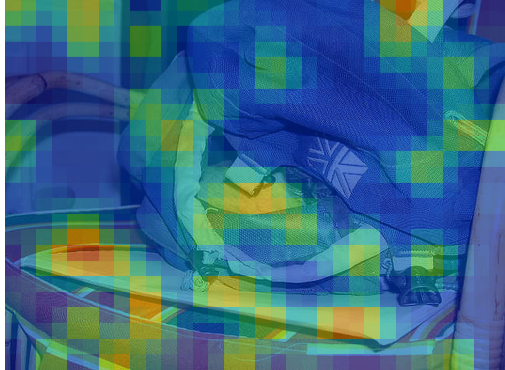} \\
\begin{minipage}[t]{\linewidth}\small
  \textbf{Q:} Is the following statement true? ``The cat is in the backpack.''
\end{minipage} &
\begin{minipage}[t]{\linewidth}\centering\small No\\\xmark\end{minipage} &
\begin{minipage}[t]{\linewidth}\centering\small No\\\xmark\end{minipage} &
\begin{minipage}[t]{\linewidth}\centering\small No\\\xmark\end{minipage} &
\begin{minipage}[t]{\linewidth}\centering\small No\\\xmark\end{minipage} &
\begin{minipage}[t]{\linewidth}\centering\small Yes\\\cmark\end{minipage} \\
\midrule
\includegraphics[width=\linewidth]{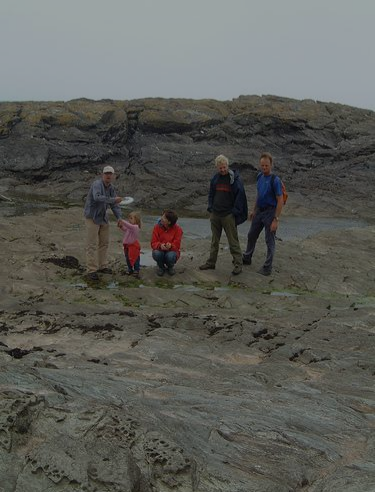} &
\includegraphics[width=\linewidth]{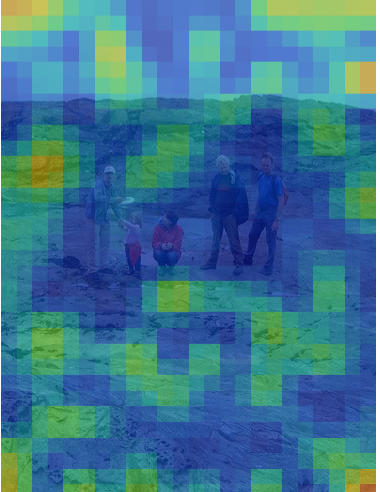} &
\includegraphics[width=\linewidth]{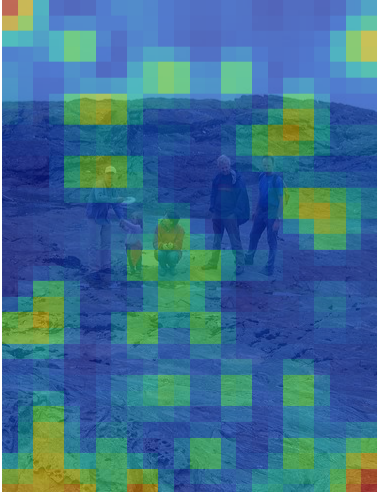} &
\includegraphics[width=\linewidth]{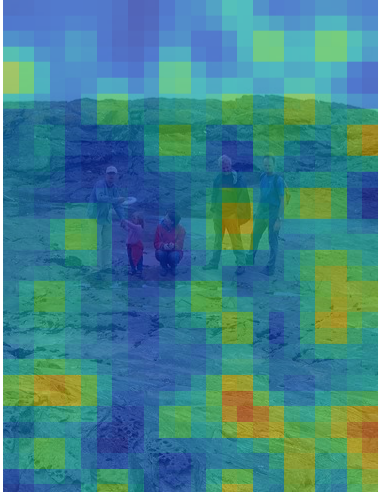} &
\includegraphics[width=\linewidth]{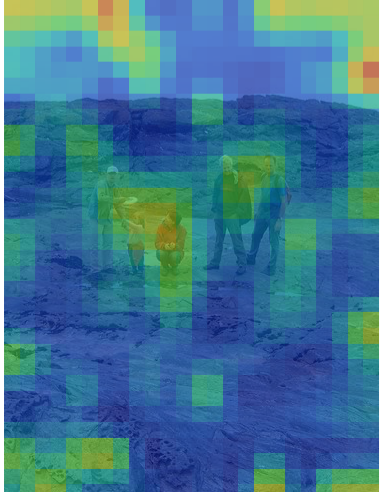} &
\includegraphics[width=\linewidth]{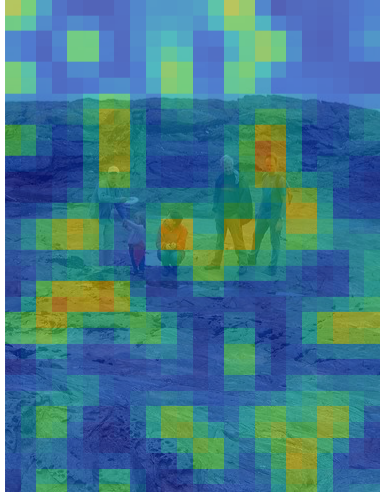} \\
\begin{minipage}[t]{\linewidth}\small
  \textbf{Q:} What color is the hat?
\end{minipage} &
\begin{minipage}[t]{\linewidth}\centering\small Black\\\xmark\end{minipage} &
\begin{minipage}[t]{\linewidth}\centering\small White\\\cmark\end{minipage} &
\begin{minipage}[t]{\linewidth}\centering\small Brown\\\xmark\end{minipage} &
\begin{minipage}[t]{\linewidth}\centering\small Brown\\\xmark\end{minipage} &
\begin{minipage}[t]{\linewidth}\centering\small White\\\cmark\end{minipage} \\
\bottomrule
\end{tabular}
}
\end{table*}

\section{Limitations}
This work focuses on visual token pruning in the prefill stage, where the computational bottleneck is most pronounced in current MLLMs and where the community has converged on well-defined baselines (VisionZip, DivPrune, CDPruner, et al.). This focus enables controlled comparison across model scales and retention ratios, a comparison that would be confounded if heterogeneous token streams were mixed. Beyond vision tokens, modalities such as audio, tool-use traces, and long textual contexts each introduce distinct temporal or structural redundancy patterns that likely call for modality-specific relevance cues and coverage estimators; we regard their systematic study as a natural and well-scoped direction for future work.




\end{document}